\documentclass[10pt,letterpaper]{article}
\usepackage[top=0.85in,left=2.75in,footskip=0.75in]{geometry}

\usepackage{amsmath,amssymb}

\usepackage{changepage}

\usepackage[utf8x]{inputenc}

\usepackage{textcomp,marvosym}

\usepackage{cite}

\usepackage{nameref,hyperref}

\usepackage[right]{lineno}

\usepackage{microtype}
\DisableLigatures[f]{encoding = *, family = * }

\usepackage[table]{xcolor}

\usepackage{array}
\usepackage{soul}
\usepackage{url}
\usepackage{array}
\usepackage{amsthm}
\usepackage{makecell}
\usepackage{longtable}
\usepackage{algorithm}
\usepackage{algorithmic}
\usepackage{graphicx}
\usepackage{diagbox}
\usepackage{colortbl}
\usepackage{xcolor}
\usepackage{titlesec}
\newcolumntype{+}{!{\vrule width 2pt}}

\newlength\savedwidth

\raggedright
\usepackage[aboveskip=1pt,labelfont=bf,labelsep=period,justification=raggedright,singlelinecheck=off]{caption}

\makeatletter
\renewcommand{\@biblabel}[1]{\quad#1.}
\makeatother

\date{}

\usepackage{lastpage,fancyhdr}
\usepackage{epstopdf}
\fancyheadoffset[L]{2.25in}
\fancyfootoffset[L]{2.25in}
\graphicspath{{pics/}{}}
\DeclareGraphicsExtensions{.jpg,.pdf,.mps,.png}

\def\diff{\mathop{\rm diff}}

\def\RR{\mathbb R}

\newcommand{\bx}{\mathbf{x}}
\newcommand{\by}{\mathbf{y}}
\newcommand{\bu}{\mathbf{u}}

\newcommand{\bbeta}{{\boldsymbol{\beta}}}

\begin{document}
\vspace*{0.2in}

\begin{flushleft}
{\Large
\textbf\newline{Are screening methods useful in feature selection? An empirical study} 
}
\newline
\\
Mingyuan Wang\textsuperscript{1},
Adrian Barbu\textsuperscript{1}
\\
\bigskip
\textbf{1} Statistics Department, Florida State University, Tallahassee, Florida, U.S.A
\\
\bigskip

%
%





* abarbu@stat.fsu.edu
* mw15m@my.fsu.edu

\end{flushleft}
\section*{Abstract}
Filter or screening methods are often used as a preprocessing step for reducing the number of variables used by a learning algorithm in obtaining a classification or regression model.
While there are many such filter methods, there is a need for an objective evaluation of these methods. Such an evaluation is needed to compare them with each other and also to answer whether they are at all useful, or a  learning algorithm could do a better job without them.
For this purpose, many popular screening methods are partnered in this paper with three regression learners and five classification learners and evaluated on ten real datasets to obtain accuracy criteria such as R-square and area under the ROC curve (AUC). 
The obtained results are compared through curve plots and comparison tables in order to find out whether screening methods help improve the performance of learning algorithms and how they fare with each other. 
Our findings revealed that the screening methods were useful in improving the prediction of the best learner on two regression and two classification datasets out of the ten datasets evaluated.



\section*{Introduction}

For the past few decades, with the rapid development of online social platforms and information collection technology, the concept of big data grew from a novel terminology in the past to one of the most powerful resources in present day. 
Especially in recent years, the sample sizes and feature space dimensions of datasets rose to levels beyond precedent. 
This development poses great challenges for machine learning in extracting the relevant variables and in building accurate predictive models on such large datasets. 

One of the most popular machine learning tasks is feature selection, which consists of extracting meaningful features (variables) from the data, with the goal of obtaining better prediction on unseen data, or obtaining better insight on the underlying mechanisms driving the response.
 
Feature selection methods have grown into a large family nowadays, with T-score\cite{ttestdavis1986statistics}, Mutual Information \cite{mutuallewis1992feature}, Relief \cite{reliefkira1992feature}, Lasso \cite{tibshirani1996regression}, and MRMR \cite{mrmrhan2005minimum}  as some of the more popular examples. 

There are three categories of feature selection methods: screening methods (a.k.a. filter methods), wrapper methods and embedded methods. 
Screening methods are independent of the model learned. This makes them less computational complex. 
However for the same reason, screening methods tend to ignore more complex feature traits. They generally provide the lowest improvement among the three. 
Wrapper methods use a learning algorithm (learner) to evaluate feature importance, which often leads to a better performance. 
But good performance comes with the possibility of overfitting, and a much higher computational demand. 
Embedded methods combine the feature selection and the model learning together, which generally makes them faster than the wrapper methods. 
However, the way that the feature selection and the learning are combined together makes them specific to the learning algorithm used.
The features selected by one kind of learner-specific embedded method may not useful for other kinds of learners.
In this study, we focus our attention on screening methods and would like an unbiased answer to the following questions:

\begin{itemize}
\item Do screening methods help to build good predictive models, or comparable models can be obtained without them?
\item How do the existing screening methods compare with each other in terms of predictive capabilities, which one is the best and which one is the worst?
\end{itemize}
To answer these questions, we evaluated different screening methods (three for regression and seven for classification) on ten real datasets, five for regression and five for classification.
The screening methods and the datasets will be described in the Methods section, but here we present our main findings.

The screening methods themselves cannot provide predictive models. For that purpose, different supervised learning algorithms such as SVM, 
Feature Selection with Annealing (FSA), Boosted Trees, and Naive Bayes were employed to construct the predictive models on the features selected by the screening methods.

For both regression and classification, experiments indicate that the screening methods are sometimes useful, in the sense they help obtaining better predictive models on some datasets.
The findings are summarized in the comparison tables from the Results section.

Through our comparison study, we intend to provide researchers with a clear understanding of some of the well known screening (filter) methods and their performance of handling high-dimensional real data.

\subsection*{Related Work}

The focus of this study is to examine the effect of screening (filter) methods on obtaining good predictive models on high-dimensional datasets. The recent literature contains several works that compare feature selection methods, including screening methods.

A recent feature selection survey \cite{li2016feature} from Arizona State University (ASU) shows a comprehensive feature selection contents, studying feature selection methods from different data type perspectives. 
The survey is very broad, examining both supervised and unsupervised learning using binary and multi-class data, whereas our study focuses on supervised learning on  regression and binary classification problems. 
The ASU study evaluates many classification datasets, but it does not have our goal of comparing feature screening methods and testing whether they are useful in practice or not. 
In this respect, we found some issues with the ASU study and we corrected them in this paper.
First, the ASU study uses the misclassification error as a measure of the predictive capability of a classifier. The misclassification error is sensitive to the choice of threshold, and is a more noisy measure than the AUC (area under the ROC curve). In our work we used the AUC instead, and obtained  performance curves that have less noise, as it will be seen in experiments. 
Second, the ASU study obtains the results with 10-fold cross-validation, and are not averaged over multiple independent runs. In our work we used 7 independent runs of 7-fold cross-validation to further increase the power of our statistical tests. 
Third, we draw our comparisons and conclusions using statistical methods based on paired $t$-tests   to obtain groups of similarly performing methods.
An earlier version of the ASU report is \cite{tang2014feature}, which is an overview of different types of feature selection methods for classification. 

In \cite{chandrashekar2014survey} are evaluated feature selection methods for flat features including filter methods, wrapper methods and embedded methods. However, tests are only conducted on low-dimensional datasets. 
In contrast we evaluate the filter methods on high dimensional datasets with 500-20,000 features and in many instances with more features than observations. 
Moreover, our goal is to compare filter methods themselves, not the filter-learning algorithm combination, since different datasets could have different algorithms that are appropriate (e.g. linear vs nonlinear). 
We achieve this goal by employing many learning algorithms and choosing the best one for each filter method and each dataset.

A comprehensive overview of the feature selection work done in recent years is shown in \cite{jovic2015review}. 
It covers feature selection methods including filter, wrapper, embedded and hybrid methods as well as structured and streaming feature selection. 
 The article also discusses existing application of these feature selection methods in different fields such as text mining, image processing, industry and bioinformatics.

Recently \cite{cai2018feature} gave another detailed and broad overview of feature selection methods. 
The authors conducted their studies of many categories of feature selection methods, including but not limited to supervised, unsupervised, semi-supervised, online learning and deep learning. 
An experiment involving five feature selection methods was conducted on classification data. All five methods are either filter or wrapper methods. 
However they conducted their experiments on only two datasets, and the didn't consider the performance of the learning algorithms without any feature selection as a comparison baseline. Therefore the paper fails to show how much the feature selection methods could improve accuracy or whether they improved accuracy at all.

From a very interesting and unique standing point, \cite{li2017recent} is an overview that focuses on the challenges currently facing feature selection research. 
They propose some solutions while at the same time reviewing existing feature selection methods. In \cite{urbanowicz2018benchmarking} is evaluated the existing Relief method and some of its variants. 
The authors implemented and expanded these methods in an open source framework called ReBATE (Relief-Based Algorithm Training Environment). 
They described these methods in great detail and conducted simulation experiments with prior knowledge of the true features. 
They used a very vast simulated data pool with many varieties. The Relief variants were also compared with three other filter methods, using as performance  measure the rate of detection of the true features. 
However, the paper didn't show if these methods can improve the performance of machine learning algorithms or if the improvement persists on real data.

Two other studies of feature selection methods are \cite{alelyani2013feature} and \cite{talavera2005evaluation}. 
In contrast to our study, they solely focus on unsupervised learning. 

With the development of feature selection research, some well written feature selection software frameworks were also introduced. FeatureSelect \cite{masoudi2019featureselect} is a newly introduced such framework, which evaluated multiple trending feature selection methods on eight real datasets. Results were compared using various statistical measures such as accuracy, precision, false positive rate and sensitivity. Their studies also evaluated five filter methods. Because the experiment didn't have learning algorithms without feature selection method as a benchmark, it again fails to show if using feature selection methods is better than not using them on these datasets. IFeature\cite{chen2018ifeature} is another feature selection software framework dedicated to Python.

Some earlier studies also exist in this field (Guyon and Elisseeff, 2003\cite{guyon2003introduction}, Sanchez-Marono et al.,2007 \cite{sanchez2007filter}, Saeys et al.,2007\cite{saeys2007review}).

\section*{Methods}
Experiments were conducted separately for regression and classification. 
For regression, the screening methods were Correlation, Mutual Information \cite{mutuallewis1992feature}, and RReliefF \cite{rrelieffrobnik1997adaptation}. These screening methods were combined with learners including Feature Selection with Annealing (FSA) \cite{barbu2017FSA}, Ridge Regression, and Boosted Regression Trees. 

For classification, the screening methods were  T-score \cite{ttestdavis1986statistics}, Mutual Information \cite{mutuallewis1992feature}, Relief \cite{reliefkira1992feature}, Minimum Redundancy Maximum Relevance (MRMR) \cite{mrmrhan2005minimum}, Chi-square score \cite{liu1995chi2},  Fisher score \cite{fisherduda2012pattern}, and Gini index \cite{gini1912variability}. They were combined with learners including FSA \cite{barbu2017FSA}, Logistic Regression, Naive Bayes, SVM, and Boosted Decision Trees. 

Among these screening methods, Mutual information, Correlation, Gini index, Fisher-score, Chi-square score and T-score select features individually. 
In contrast, MRMR requires to calculate the redundancy between the already selected features and the current feature, and Relief requires to calculate the distance between two observations using the Euclidean norm, so that one can determine the nearest neighbor with the same label and with a different label. The calculation of the Euclidean norm involves all the feature values. Consequently, these two methods select features in combination and are slower than the other methods.

\subsection*{Evaluation of Screening Methods}
The predictors of all datasets were normalized to zero mean and standard deviation 1 in a pre-processing step.
For each dataset, experimental results were obtained as the average of 7 independent runs of 7-fold cross-validation.
For each run, a random permutation of the dataset was generated and the data was split into seven approximately equal subsets according to the permutation. Then a standard full 7-fold cross-validation was performed as follows. Each fold consists of testing on one of the subsets after training on the other six. This procedure was run with each of the seven subsets as the test set and the other six as the training set. For each fold, each one of the screening methods mentioned above was used to reduce the dimension of the feature space to the desired size, then a learning algorithm using preset parameter values was applied on the selected features to obtain the model. The predictions of the model on the test subset for each fold were combined to obtain a vector of test predictions on the entire dataset, which was used to obtain performance measures ($R^2$ for regression and AUC (Area under the ROC curve) for classification). To increase accuracy, these performance measures were averaged over seven independent runs on different permutations of the data.

To insure the consistency of the comparison, the number of features that were selected by each screening method was kept the same for each dataset. 
For each dataset (except Wikiface), 30 different values of the number of selected features were assigned. 
Plots were used to compare the average performance over the 7 cross-validated runs of different combinations of screening method and learner. 
Also for each combination, the optimal number of selected features was selected based on the maximum average test performance over the 7 cross-validated runs. Pairwise t-tests at the significance level $\alpha$=0.05 were used to compare between different combinations to see if they are significantly different.

\subsection*{Construction of the Tables of Groups}
Groups of screening method-learner combinations that are not significantly different from each other were constructed as follows (we use paired $t$-tests to obtain $p$-values when comparing different methods combinations and set 0.05 as the significance level in our experiment). 
The screening method-learner combinations are first sorted in descending order of their peak performance. 
Then starting from the first combination F downward, the last combination in the sequence that is not statistically significantly different from combination F is marked as combination L. 
All combinations between F and L are put into the same group. 
The same procedure was used for other combinations along the sequence. All these tables of groups are provided in the Supporting Information.

\subsection*{Construction of the Comparison Tables }
Comparison tables were established based on how many times each screening method-learner combination  appeared in the group tables. 
Three kinds of counting methods were applied. 

\noindent 1) {\em The number of datasets where the screening method performed significantly better than no screening} for different learning algorithms. For each learning algorithm, it is the number of datasets on which the screening method appeared in higher group tiers than the same learning algorithm without screening. 
This counting method is used to construct Table \ref{tab:regFilterB} and Table \ref{tab:clFilterB} except the ``Best algorithm'' column.

\noindent 2) {\em The number of datasets where the screening method was significantly better than the best performing algorithm with no screening (usefulness per dataset)}.  For each dataset, we checked for each screening method whether it appeared with a learning algorithm in a higher group tier than the best learning algorithm without screening. 
This counting method is used to construct Table \ref{tab:regFilterT} and Table \ref{tab:clFilterT} and the ``Best algorithm'' column of Tables \ref{tab:regFilterB} and \ref{tab:clFilterB}.
The column named ``Total Count'' is generated for each screening method from the sum of the counts across all datasets. 

\noindent 3)  {\em The number of datasets where each filter-learning algorithm combination was in the top performing group (top performing)}. This counting method is used in  Table \ref{tab:regA} and Table \ref{tab:clA}.
 \begin{table}[htb]
\vskip -1mm
\begin{center}
\caption{The datasets used for evaluating the screening methods. The parameter $\tau$ controls the number of selected features as $[(4t)^{\tau}], t=\overline{1, 30}$.}\label{tab:dataset}
\begin{tabular}{|l|c|c|c|c|c|}
\hline
Dataset &\thead{Learning type} &\thead{Feature type} &\thead{Number of\\ features} &\thead{Number of\\ observations} &\thead{$\tau$}\\[0.5ex]
\hline
\hline
\href{https://ani.stat.fsu.edu/~abarbu/Research/MouseBMI.zip}{Mouse BMI} \cite{wang2006genetic} &Regression &Continuous &21575 &294 &1.825\\[0.5ex]
\hline
\href{https://ani.stat.fsu.edu/~abarbu/Research/tumor.zip}{Tumor} \cite{tumorNCIGDC}&Regression &Continuous &16790 &1750 &1.825\\[0.5ex]
\hline
\href{https://archive.ics.uci.edu/ml/datasets/ujiindoorloc}{Indoorloc} \cite{torres2014ujiindoorloc}&Regression &Continuous &520 &20294 &1.25\\[0.5ex]
\hline
\href{https://ani.stat.fsu.edu/~abarbu/Research/WikiFace.zip}{Wikiface} \cite{Rothe-IJCV-2016} &Regression &Continuous &4096 &53040 &1.65\\[0.5ex]
\hline
\href{http://www.coepra.org/CoEPrA-2006/CoEPrA-2006_Regression_003.zip}{CoEPrA2006} \cite{CoEPrA20063} &Regression &Continuous &5787 &133&1.68\\[0.5ex]
\hline
\href{http://featureselection.asu.edu/datasets.php}{Gisette} \cite{guyon2005result} &Binary Classification &Continuous &5000 &7000 &1.73\\[0.5ex]
\hline
\href{https://archive.ics.uci.edu/ml/datasets/dexter}{Dexter} \cite{guyon2005result} &Binary Classification &Continuous &20000 &600 &1.78\\[0.5ex]
\hline
\href{http://featureselection.asu.edu/datasets.php}{Madelon} \cite{guyon2005result} &Binary Classification &Continuous &500 &2600 &1.25\\[0.5ex]
\hline
\href{http://featureselection.asu.edu/datasets.php}{SMK\_CAN\_187} \cite{spira2007airway} &Binary Classification &Continuous &19993 &187 &1.78\\[0.5ex]
\hline
\href{http://featureselection.asu.edu/datasets.php}{GLI\_85} \cite{freije2004gene} &Binary Classification &Continuous &22283 &85 &1.78\\[0.5ex]
\hline
\end{tabular}
\end{center}
\vskip -6mm
\end{table}
\section*{Results}
\subsection*{Data sets}
Five datasets were used for regression and five datasets for classification, with the specific dataset details given in Table \ref{tab:dataset}.

The regression dataset Indoorloc is available on the UCI Machine Learning Repository \cite{uciML}. The original dataset has eight indicator columns including longitude, latitude and so on. In our study, we only used the latitude as response. We combined the training and validation data files and deleted all duplicated observations due to the removing of the other seven indicator columns. The dataset Tumor was extracted from TCGA ( The Cancer Genome Atlas). The response of this dataset is the survival time(in days) of the patient, and the predictors represent  gene expression levels.
The classification datasets Gisette, Dexter, Madelon are part of the NIPS 2003 Feature selection challenge \cite{guyon2005result} and are also available on the UCI Machine Learning Repository. 

The dataset Wikiface is a regression problem of predicting the age of a person based on the person's face image, and was obtained from the Wikiface images \cite{Rothe-IJCV-2016}.
A CNN (Convolutional Neural Network) vgg-face \cite{parkhi2015deep} pre-trained for face recognition was applied to each face and the output of the 34-th layer was used to generate a 4096 feature vector for each face. 
This 4096 dimensional vector was used as the feature vector for age regression, with the age value from the original Wikiface data as the response.

\subsection*{Regression Results}

The following results are based on the output generated using Matlab 2016b \cite{Matlab2016b}. For RReliefF, correlation score, ridge regression and boosted regression trees we used their Matlab 2016b implementation. Mutual information for regression was implemented by ourselves. For FSA we used the Github\footnote{\url{https://github.com/barbua/FSA}} implementation from its original authors.

\begin{figure}[htb]
\vspace{-4mm}
\centering
\hspace{-4mm}
\includegraphics[width=0.5\linewidth]{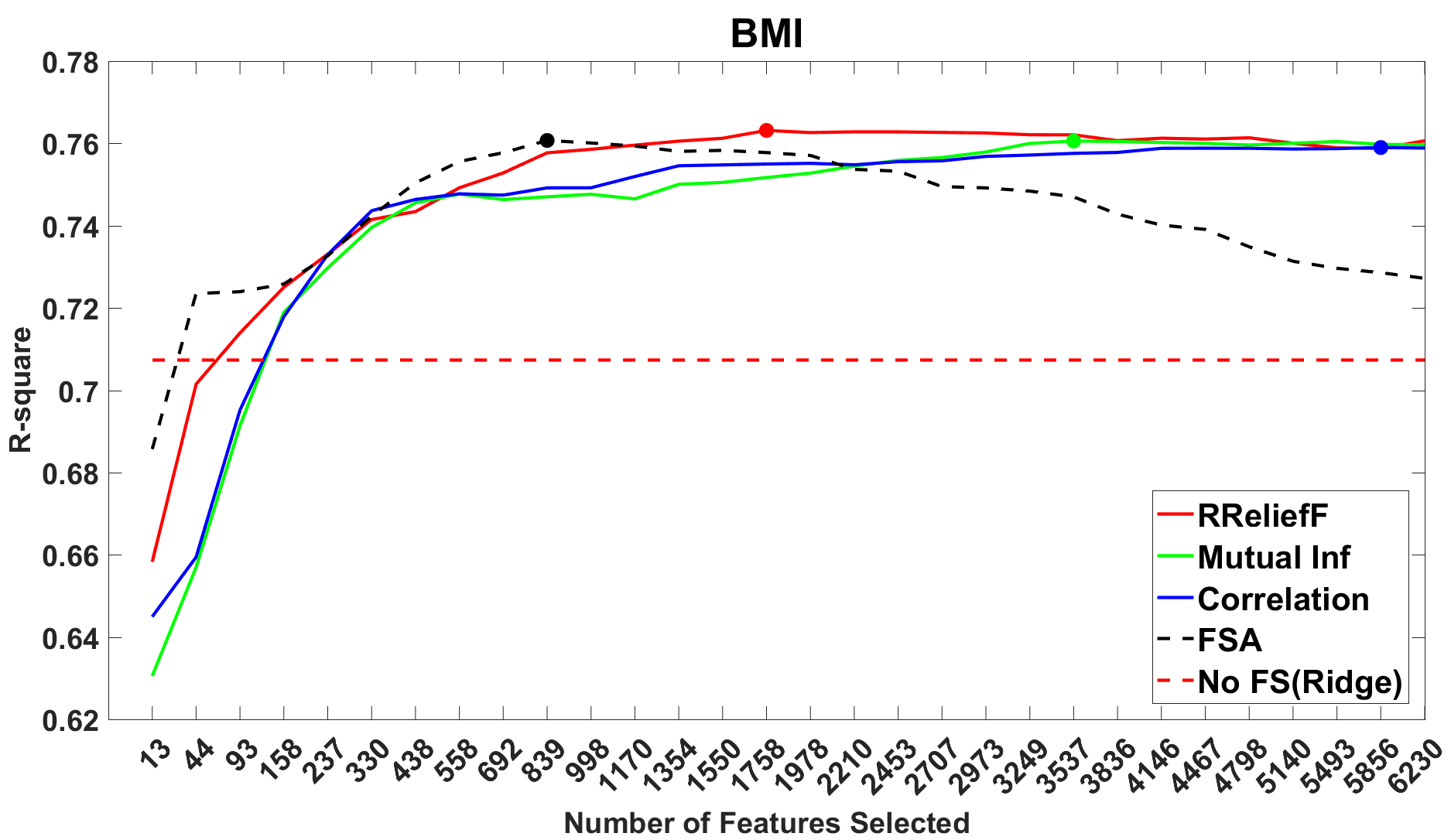}
\includegraphics[width=0.5\linewidth]{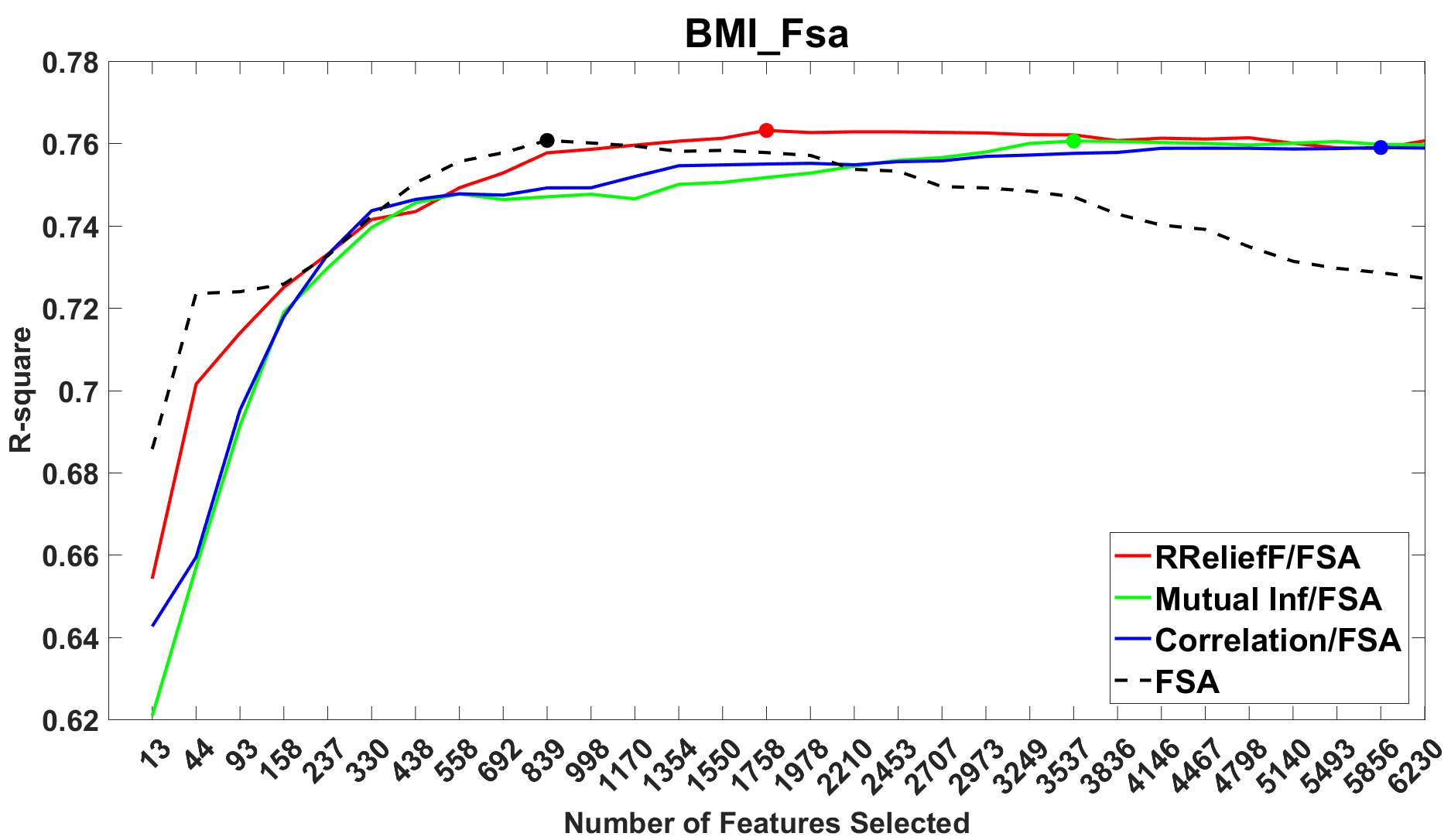}\\
\hspace{-4mm}
\includegraphics[width=0.5\linewidth]{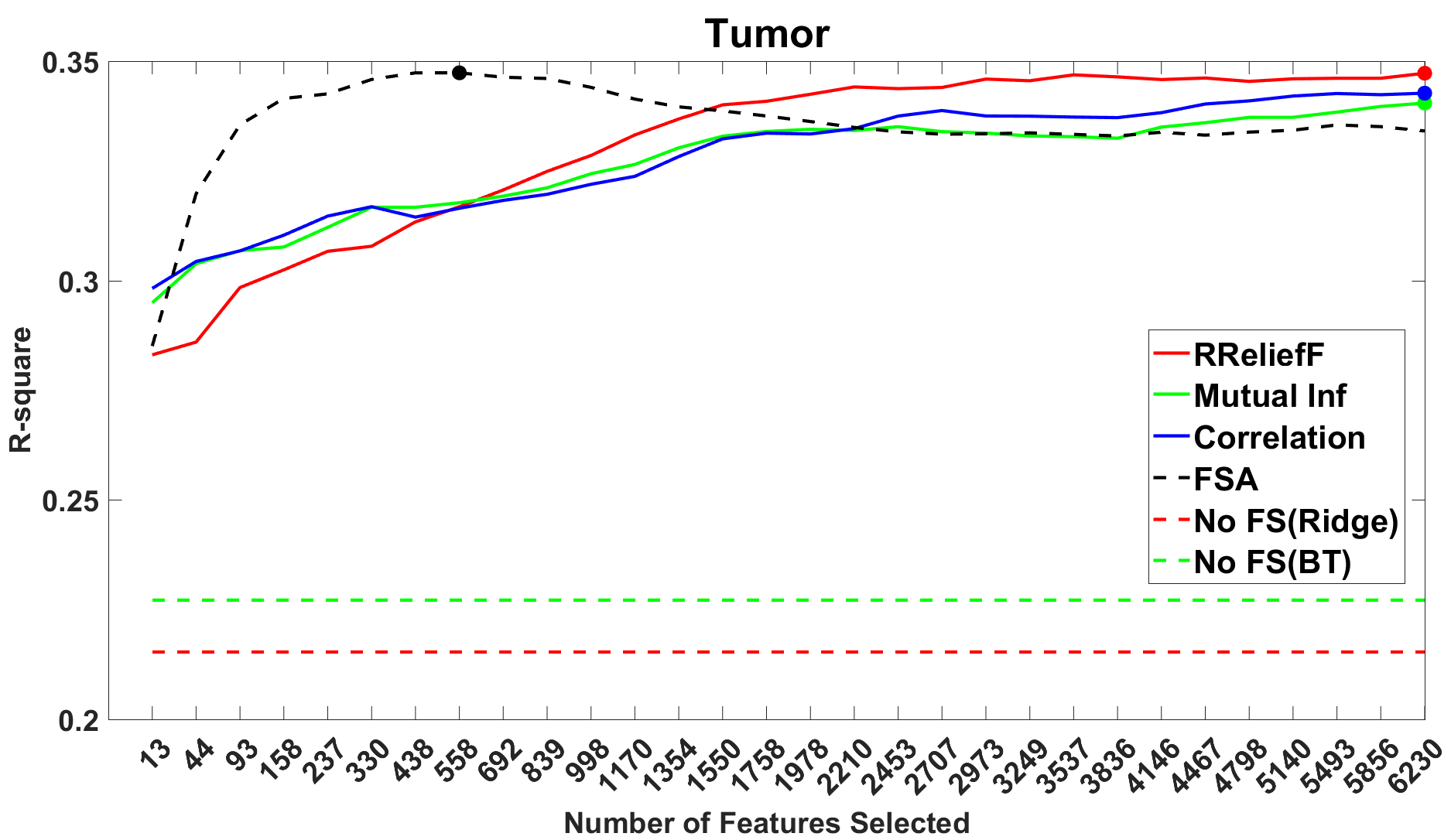}
\includegraphics[width=0.5\linewidth]{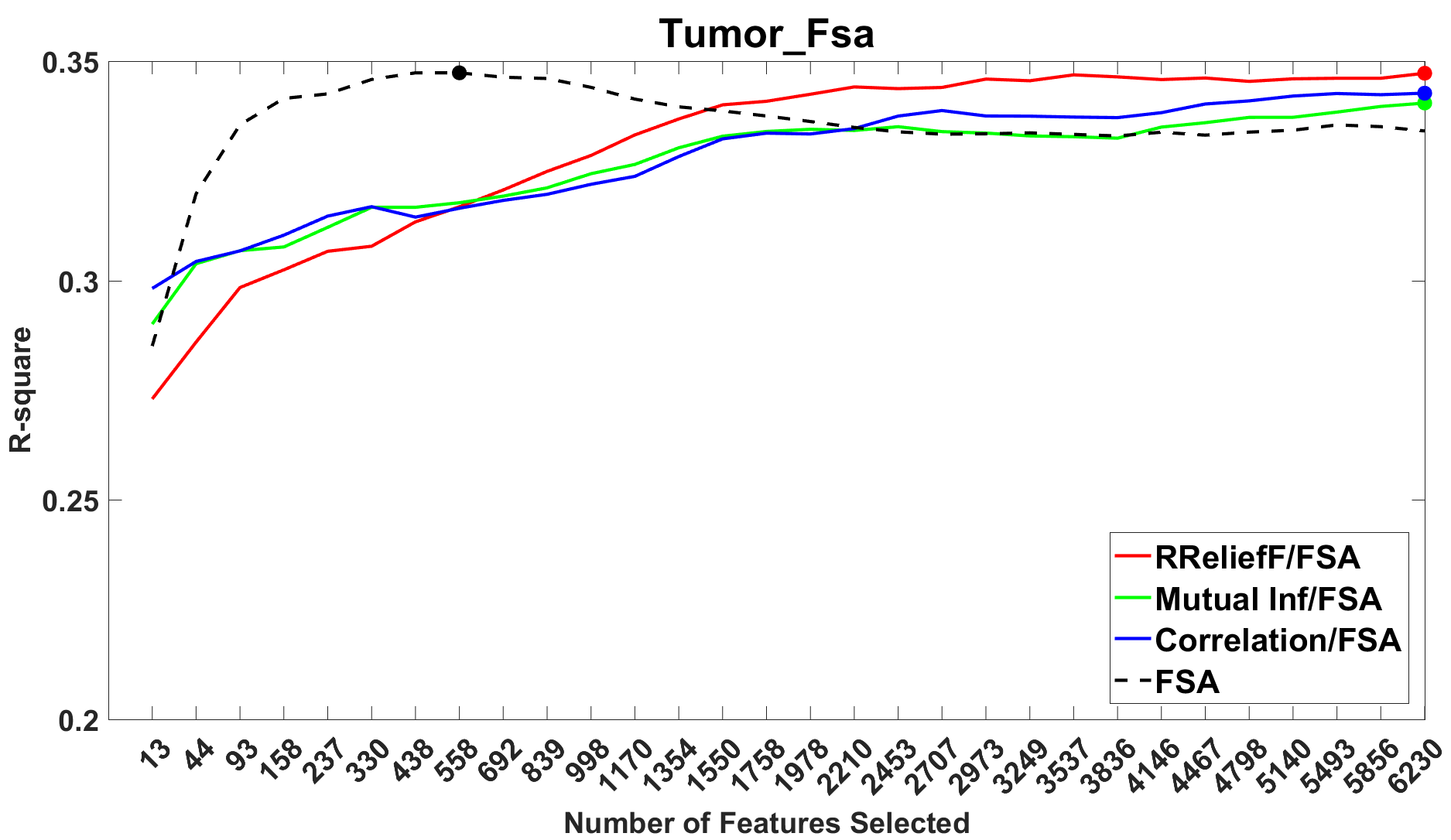}
\caption{Performance plots of methods with and without feature screening. Left: for each screening method are shown the maximum $R^2$ value across all learners. Right: $R^2$ of the screening methods with the best learner for this data (FSA).}\label{fig:BMI}
\vspace{-6mm}
\end{figure}
\subsubsection*{Performance Plots}

\begin{figure}[t]
\centering
\hspace{-4mm}
\includegraphics[width=0.5\linewidth]{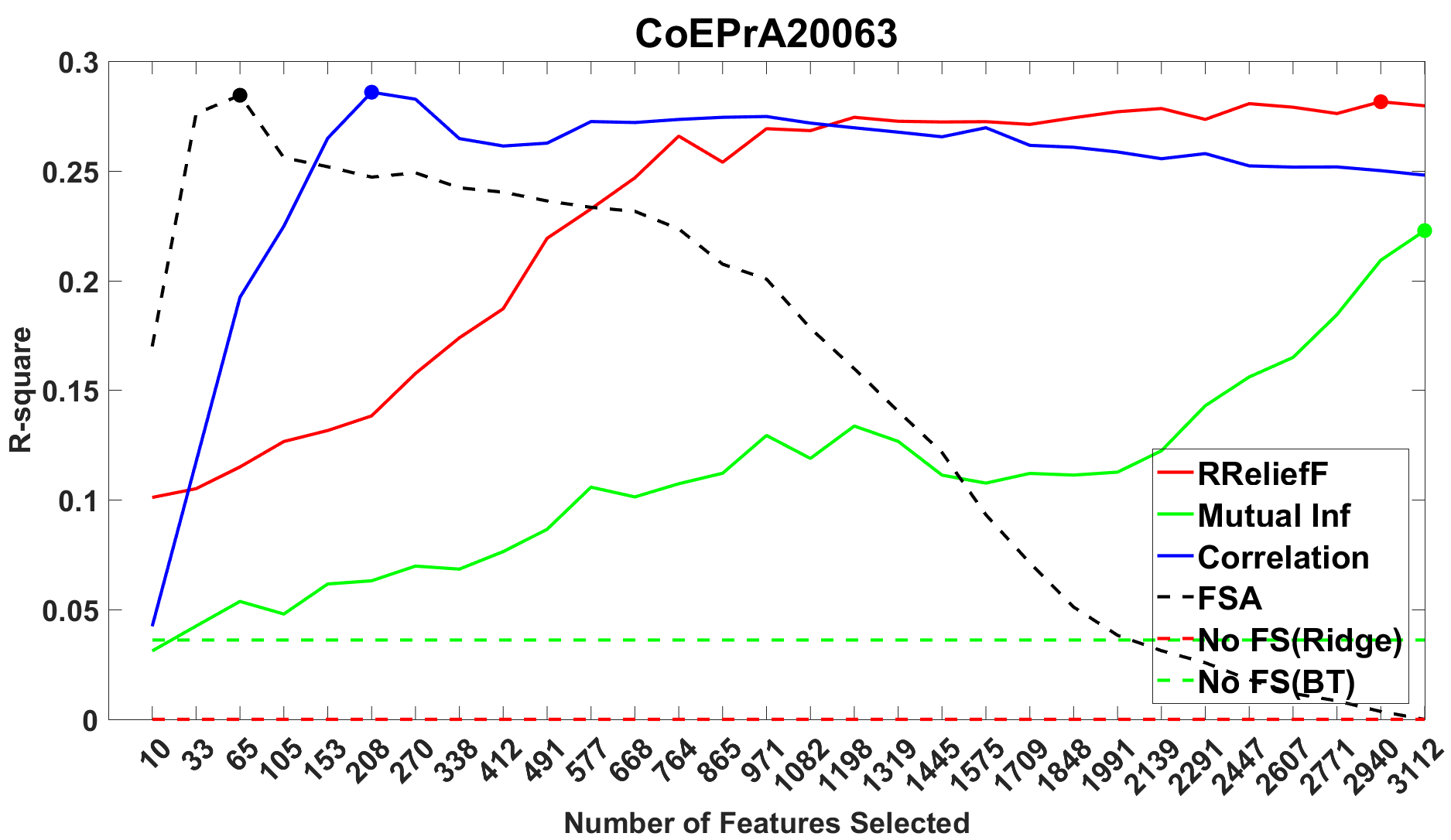}
\includegraphics[width=0.5\linewidth]{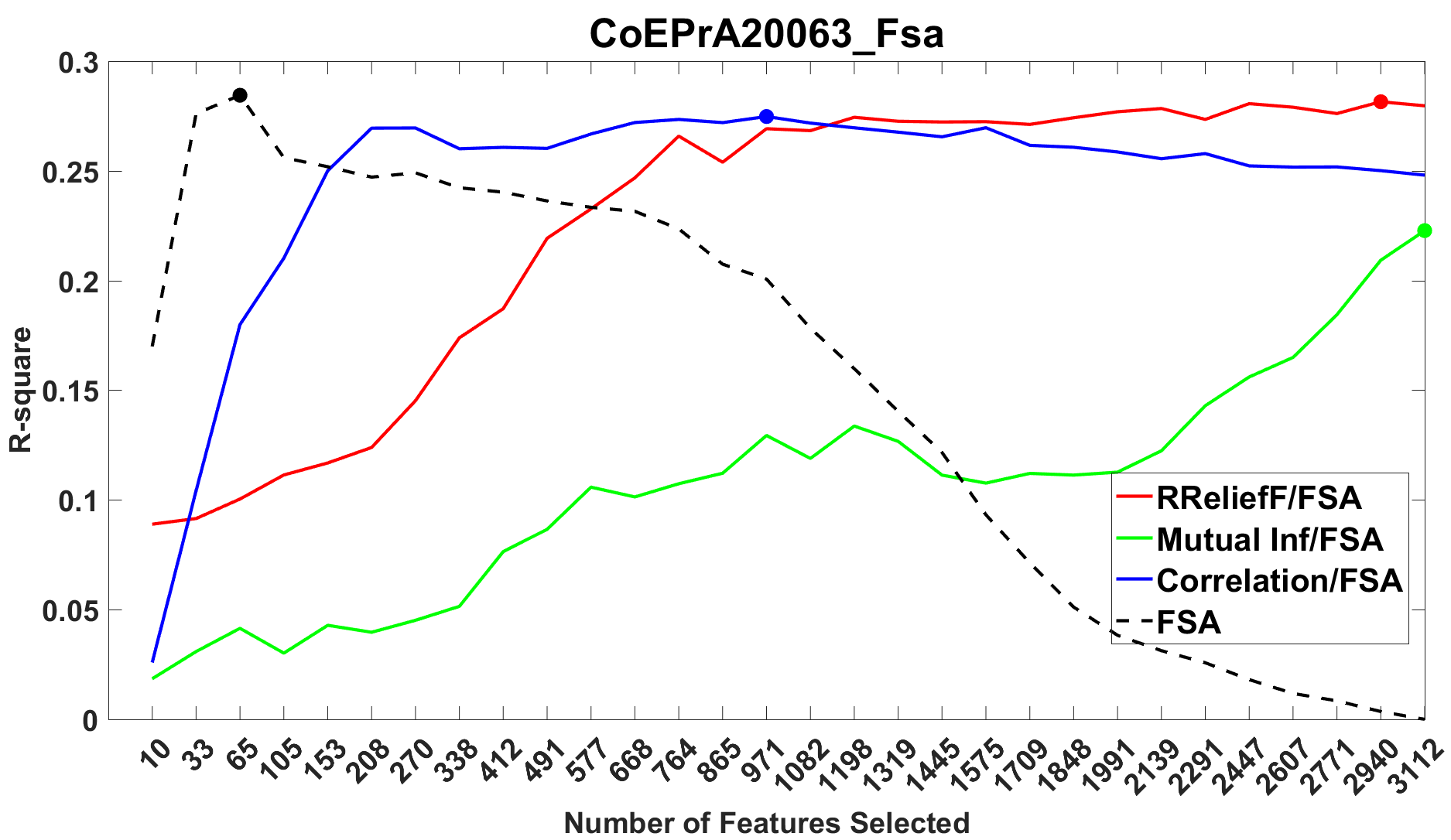}\\
\hspace{-4mm}\includegraphics[width=0.5\linewidth]{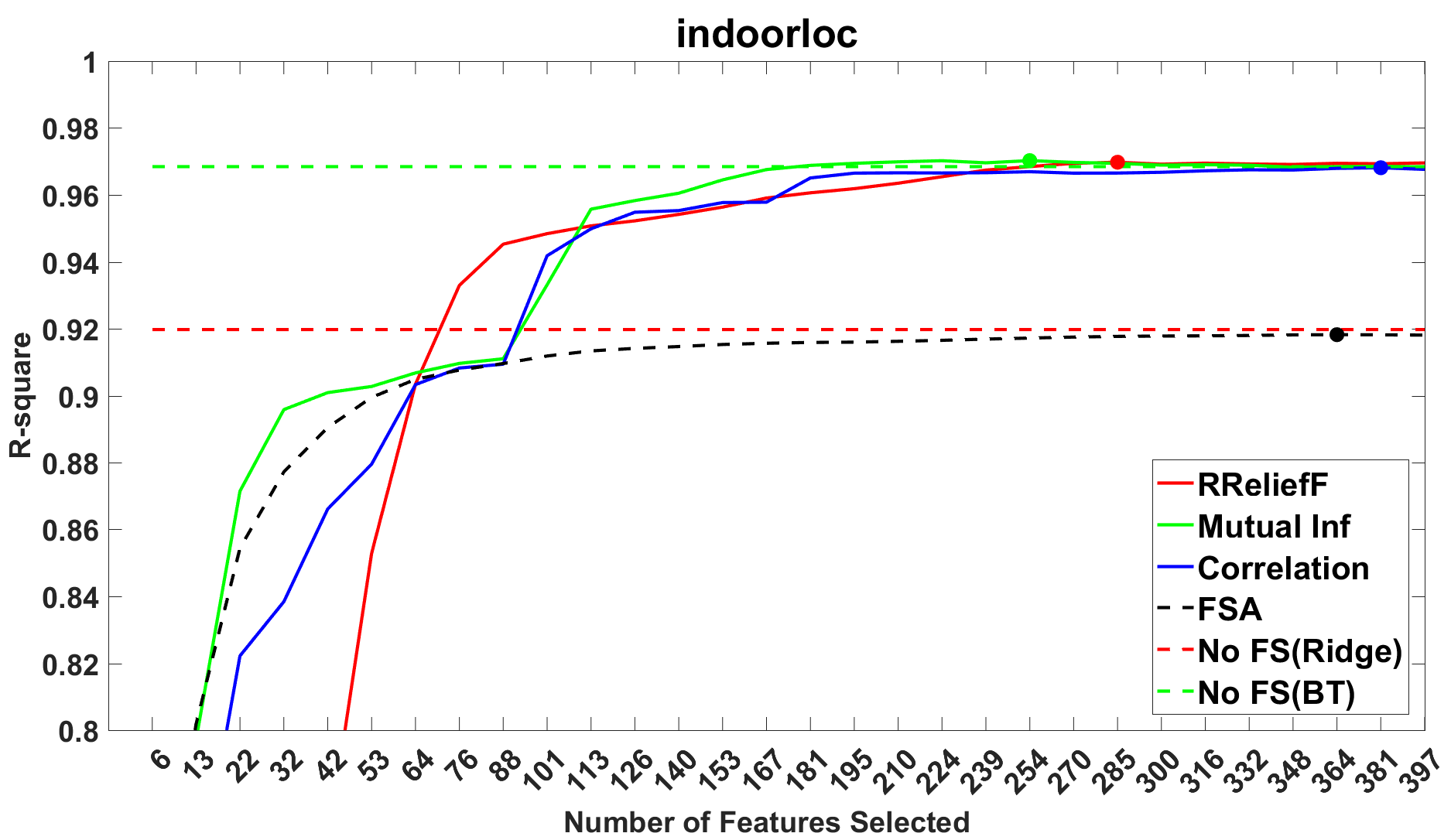}
\includegraphics[width=0.5\linewidth]{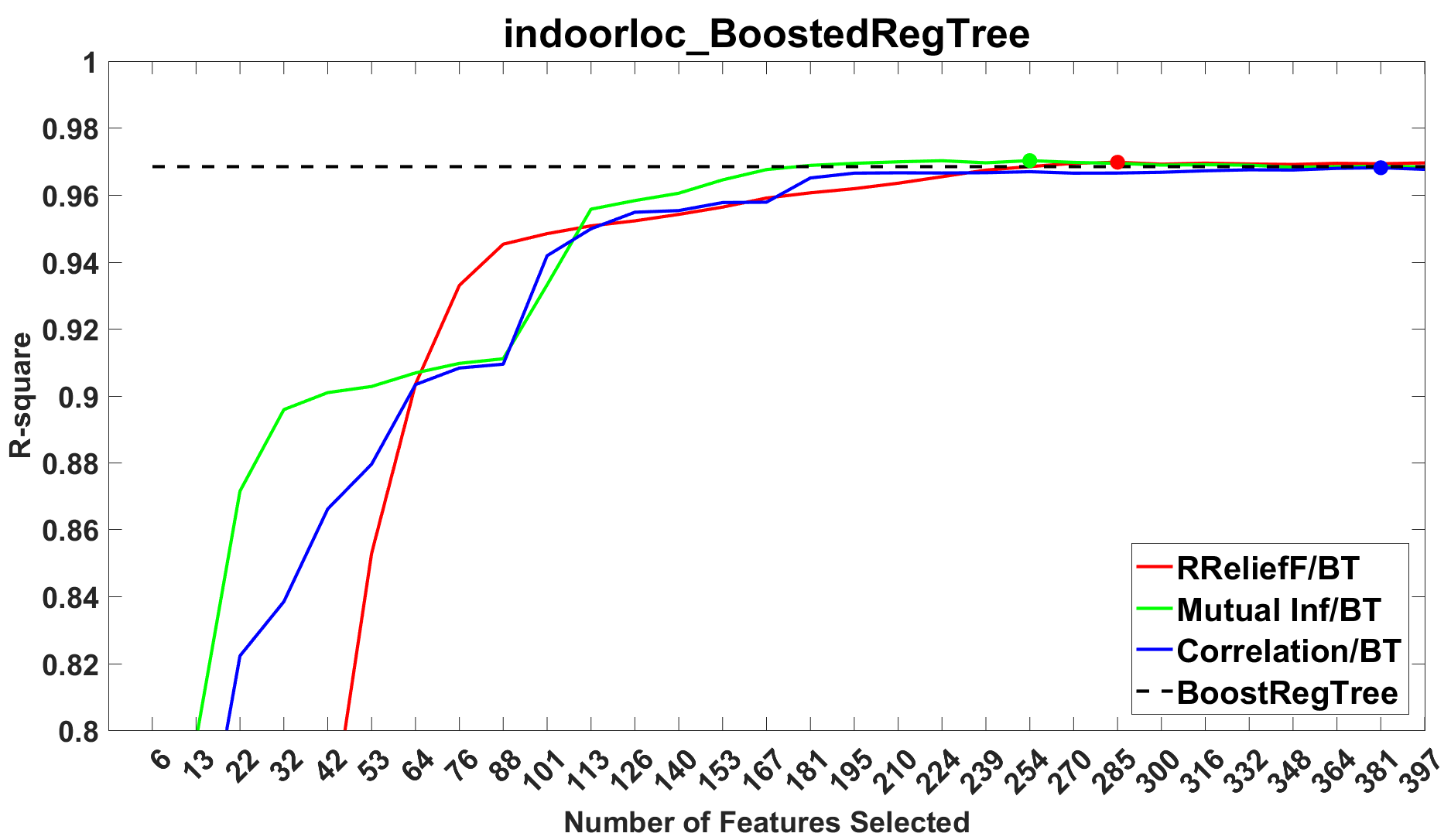}\\
\hspace{-4mm}\includegraphics[width=0.5\linewidth]{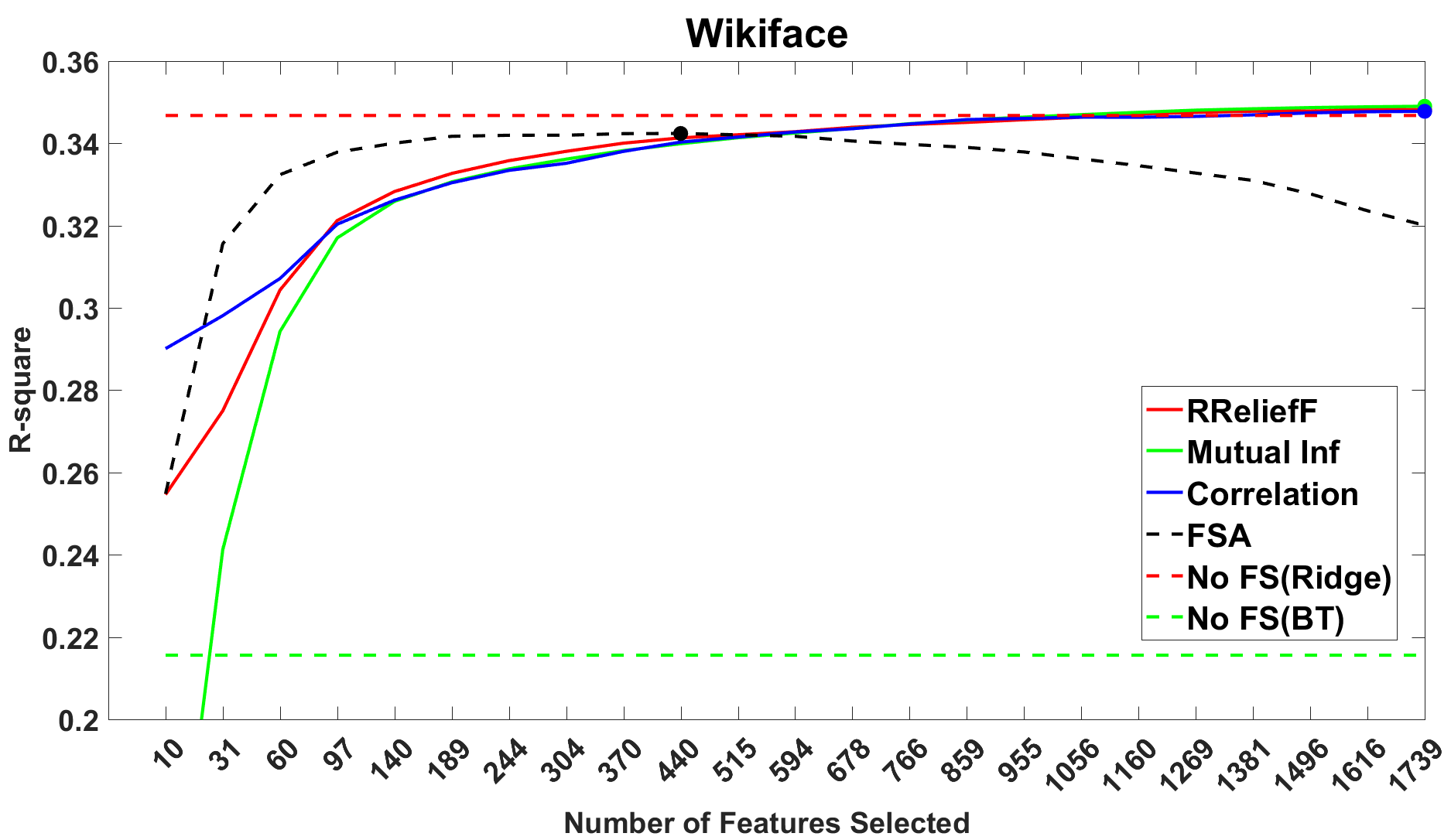}
\includegraphics[width=0.5\linewidth]{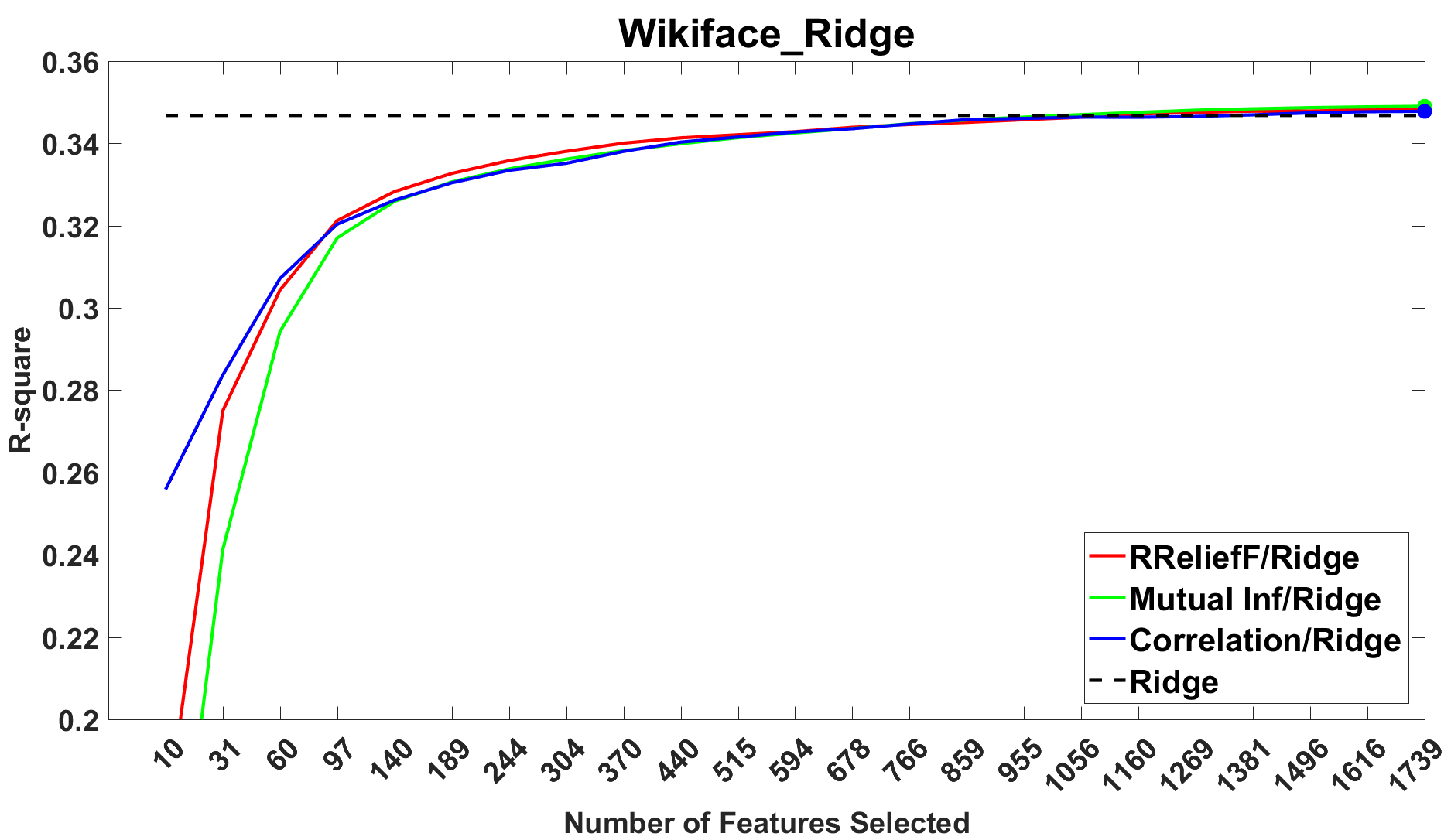}
\caption{Performance plots of methods with and without feature screening. Left: for each screening method are shown the maximum $R^2$ value across all learners. Right: $R^2$ of the screening methods with the best learner for this data (ridge).}\label{fig:coepra}
\vspace{-6mm}
\end{figure}

For the regression datasets, the plots from Fig. \ref{fig:BMI} and \ref{fig:coepra} show the $R^2$ value vs. the number $M_i$ of selected features, where $M_i=[(4i)^{\tau}], i=1,...,30$. The value of $\tau$ for each dataset is given in Table \ref{tab:dataset}.


In Fig \ref{fig:BMI}, left, are shown the $R^2$ of the best learning algorithm vs. the number of features selected by a screening method for the BMI and tumor datasets. 
Observe that these datasets are both gene expression datasets with many features and few observations. 
In Fig \ref{fig:BMI}, right, are shown the $R^2$ of FSA (the best overall learning algorithm) vs. the number of features selected by a screening method.
Except a slightly higher value given by RReliefF on the BMI data, overall the screening methods did not show higher scores than that of the optimal regression learners for the BMI and Tumor datasets. 
The plots on the right show that the screening methods even needed to select more features to obtain similar performance to FSA without screening.

In Fig \ref{fig:coepra}, left, are shown the $R^2$ of the best learning algorithm vs. the number of features selected by a screening method for the other three regression datasets. 
In Fig \ref{fig:coepra}, right, are shown the $R^2$ of the best overall learning algorithm in each case (ridge for CoEPrA and Wikiface, boosted trees for Indoorloc) vs. the number of features selected by a screening method. 
From the plots we observe that screening methods give slightly better results than the learning algorithms without screening on the Indoorloc and Wikiface datasets. 
The statistical significance of the improvement can be seen in the table of groups from the supporting information or in the comparison tables below.


\subsubsection*{Comparison Tables}
The counts in the comparison table are based on the table of groups from the supporting information.

\begin{table}[htb]
\vspace{-2mm}
\begin{center}
\caption{Overview of the number of datasets where each feature screening method performed significantly better than no screening for different learning algorithms and than the best performing algorithm (larger numbers are better).}
\label{tab:regFilterB}
\begin{tabular}{|l|c|c|c|c|}
\hline
{Screening Method} &FSA &Ridge &Boost Tree &Best algorithm \\
\hline
\hline
{RReliefF \cite{rrelieffrobnik1997adaptation}} &0 &4& 3&2\\
\hline
{Mutual Information \cite{mutuallewis1992feature}} &0 &3&3& 2\\
\hline
{Correlation} &0 & 4& 3& 1\\
\hline
\end{tabular}
\end{center}
\vspace{-5mm}
\end{table}
In Table \ref{tab:regFilterB} is shown the number of datasets where a filter method helps an algorithm perform significantly better, and the number of datasets where a screening method significantly improves a learning algorithm compared to the best performing learning algorithm without screening. It is shown that screening methods have relatively good performance  with ridge regression and boosted regression trees on datasets tested. They work on 3-4 out of 5 datasets. RReliefF method and Mutual information method have slightly better performance than Correlation method when only comparing with the best learner without screening methods.
\begin{table}[t]
\vspace{-3mm}
\begin{center}
\caption{Ranking of feature screening methods for regression by number of datasets where screening method was significantly better than the best performing no screening method. (larger numbers are better)}
\label{tab:regFilterT}
\begin{tabular}{|l|c|c|c|c|c|c|}
\hline
{Screening Method} &BMI &Tumor &{CoEPrA}& Indoorloc &Wikiface&Total Count \\
\hline
\hline
{RReliefF} &=&=&=&*& *&2\\
\hline
{Mutual Information.} &=&&&*& *& 2\\
\hline
{Correlation} &=&&=&=& *& 1\\
\hline
\end{tabular}
\end{center}
\vspace{-5mm}
\end{table}

In Table \ref{tab:regFilterT} is shown a ``*'' for each dataset and each screening method when it has a learning algorithm that obtains significantly better performance than the best learning algorithm without screening. 
An ``='' sign shows for each dataset when the screening method is in the same performance group as the best learning algorithm without screening method (so it does no harm). 
We can see that Mutual Information and RReliefF worked on the Indoorloc dataset and all three screening methods worked on the Wikiface data. However, the screening methods didn't provide performance improvement on the other three regression datasets. 
It is also shown that only in few occasions that screening methods harm the best learning algorithm. This is shown by the blank cells in table.
\begin{table}[htb]
\vspace{-3mm}
\begin{center}
\caption{Number of datasets each combination was in the top performing group.}\label{tab:regA}
\begin{tabular}{|l|c|c|c|}
\hline
\backslashbox{Filter}{Learners}&FSA& Ridge& Boost Tree \\
\hline
\hline
{RReliefF}&3& 0& 1\\
\hline
{Mutual Information}& 1&1&1\\
\hline
{Correlation}& 2&1&0\\
\hline
{---}&3& 0&0\\
\hline
\end{tabular}
\end{center}
\vspace{-5mm}
\end{table}

In Table \ref{tab:regA} is shown the counts of screening method-learner combinations that are in the top group. The combination of FSA with screening methods worked on more regression datasets than the other screening-learner combinations. However it was not a significant improvement compared to FSA without screening, which also worked on 3 out of 5 datasets.
\begin{table}[h]
\begin{center}
\caption{Ranking of feature screening methods for regression by the number of times each was in the top performing group. (larger numbers are better)}
\label{tab:regFilterA}
\vskip 1mm
\begin{tabular}{|l|c|c|}
\hline
 &\multicolumn{2}{c|}{Top performing}\\ 
{Screening Method} &Method-Algorithm &Method\\
\hline
\hline
{RReliefF} &4&4\\
\hline
{Mutual Information} &3&3\\
\hline
{Correlation} &3&2\\
\hline
{No Screening} &3&3\\
\hline
\end{tabular}
\end{center}
\vspace{-6mm}
\end{table}

In Table \ref{tab:regFilterA} is shown the number of times each screening method was in the top performing group. In the first column, these methods were counted together with the learning algorithms they were applied. 
So there can be at most 15 counts (For each screening method there are three learning algorithms and  five datasets total) in each cells. 
The second column shows the counts withe the best learning algorithm for each method, so there can be at most 5 counts in each cell. 
The table shows that RReliefF has the best performance, which is larger that the worst performance (correlation) by 2. 
Among the three screening methods only RReliefF has a higher count than non screening.
\begin{figure}[t]
\vspace {-4mm}
\centering
\hspace{-4mm}
\includegraphics[width=0.5\linewidth]{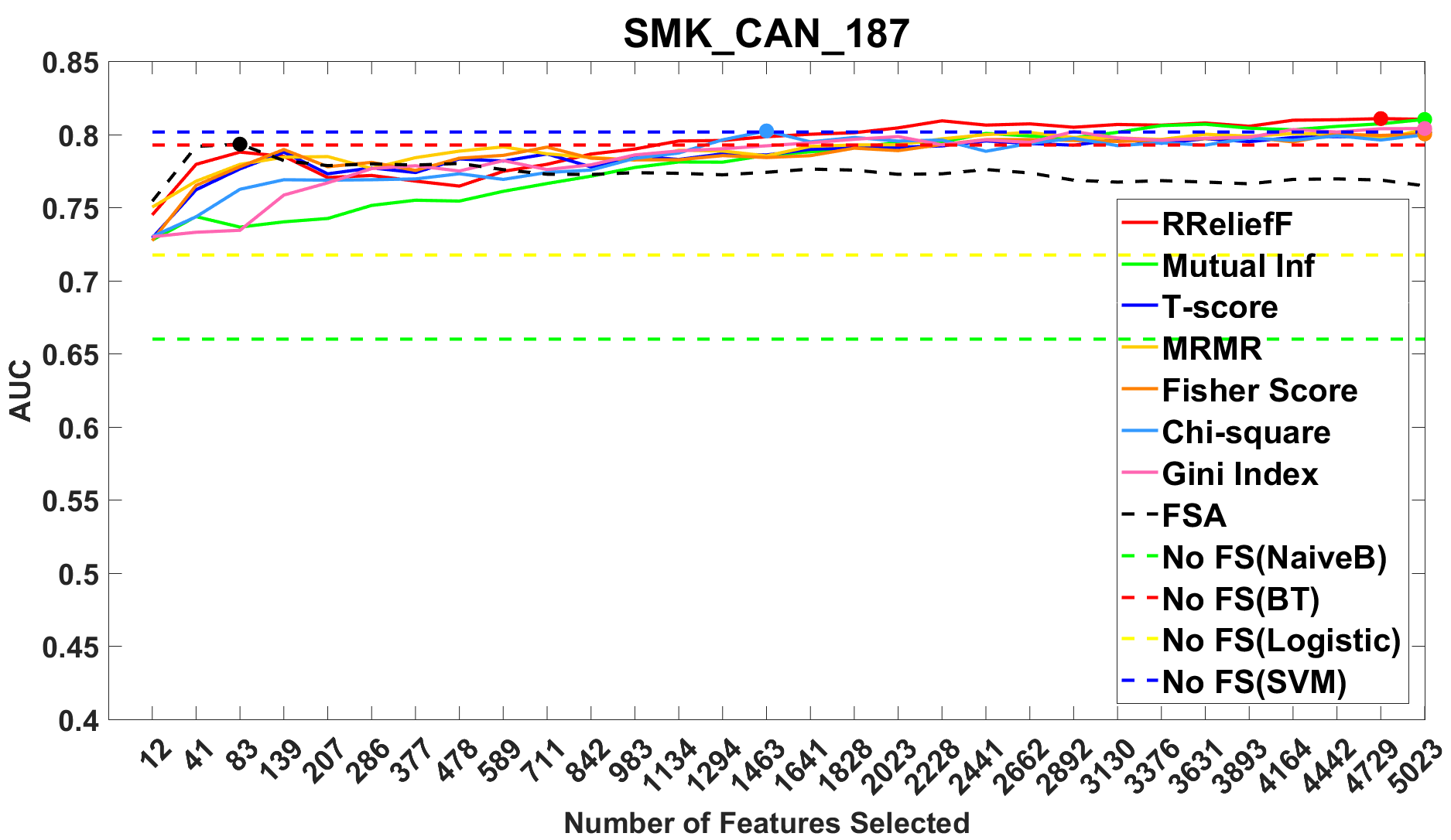}
\includegraphics[width=0.5\linewidth]{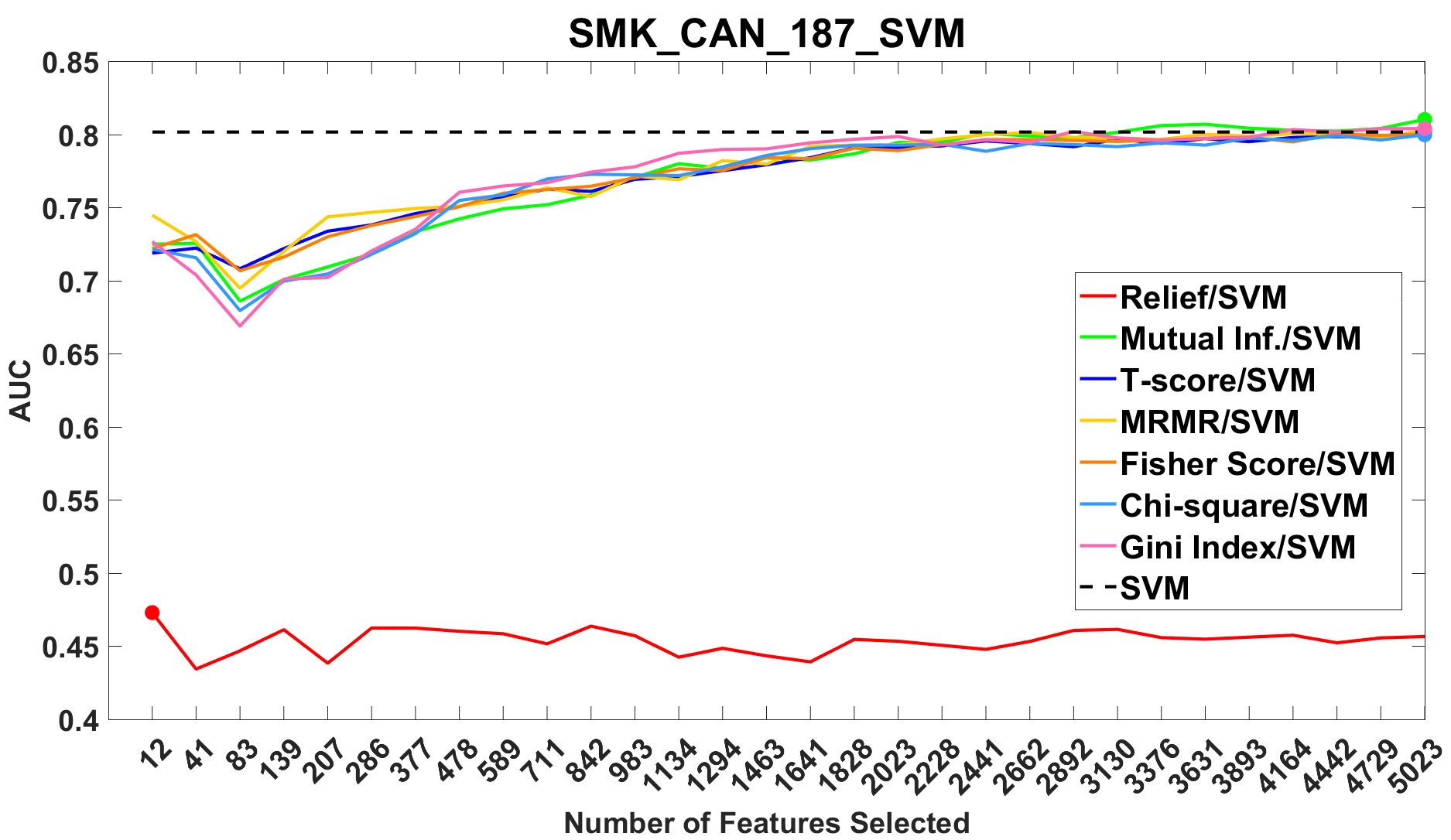}\\
\vspace {-2.5mm}
\hspace{-4mm}
\includegraphics[width=0.5\linewidth]{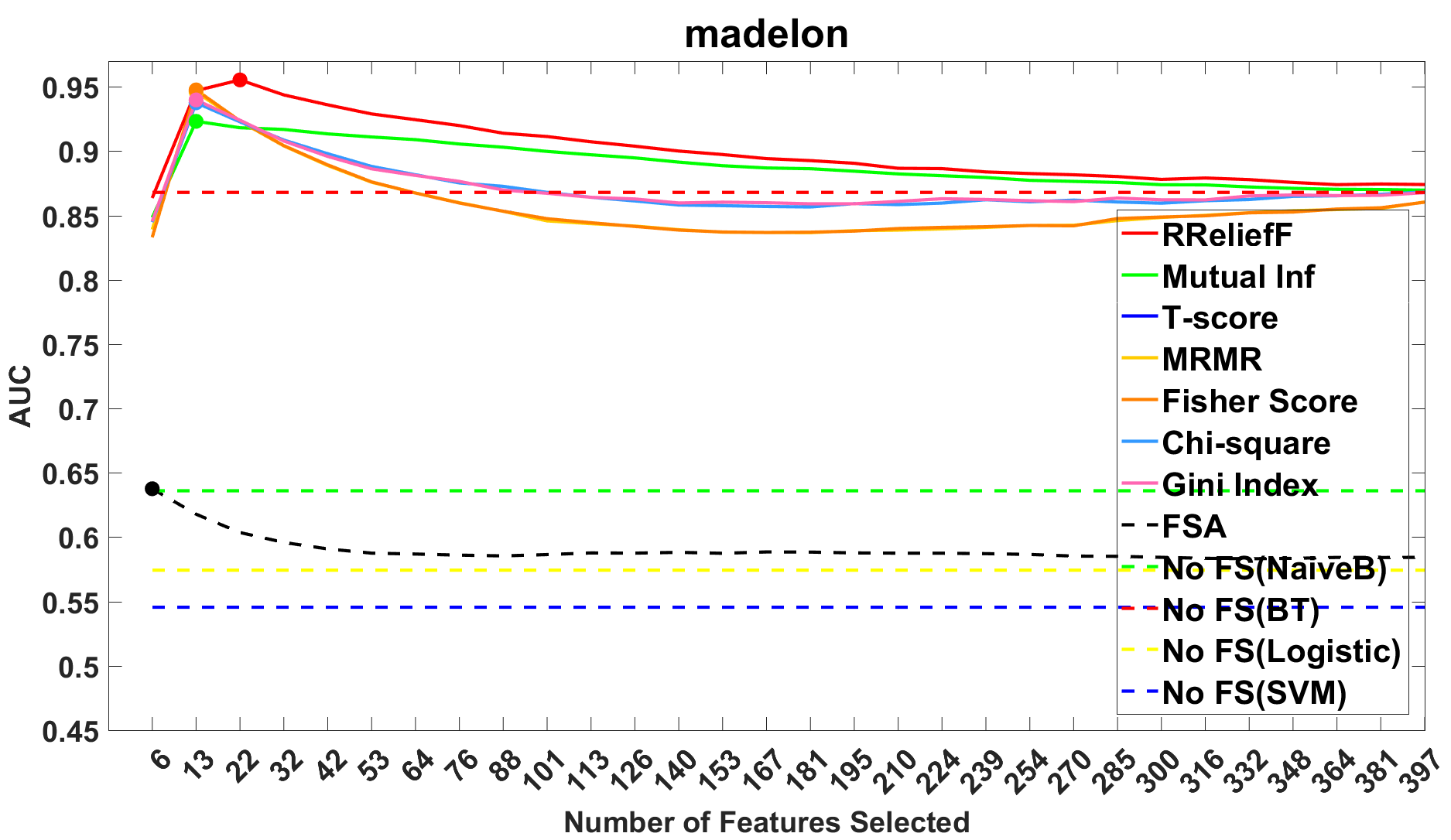}
\includegraphics[width=0.5\linewidth]{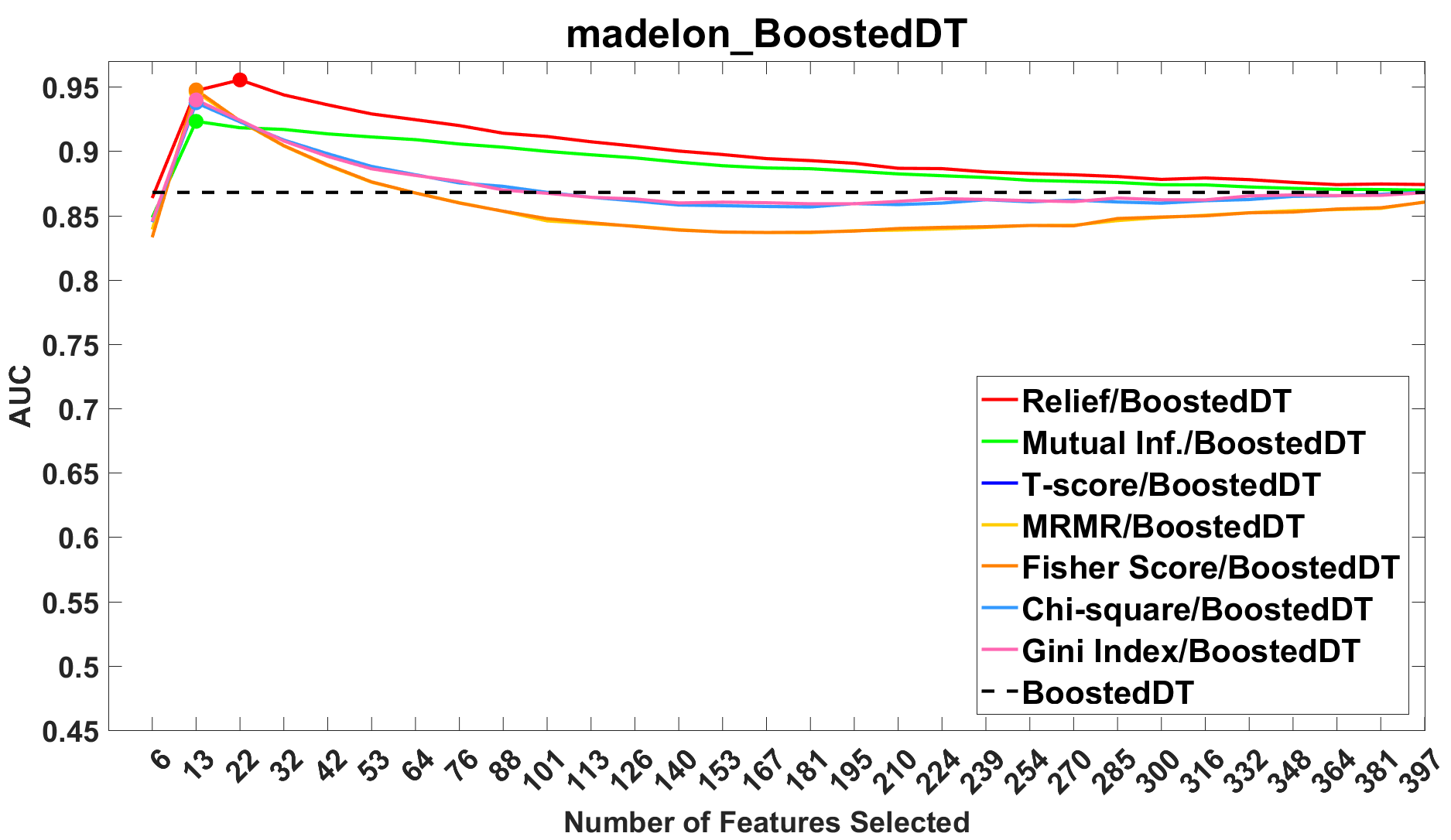}\\
\vspace {-2.5mm}
\hspace{-4mm}
\includegraphics[width=0.5\linewidth]{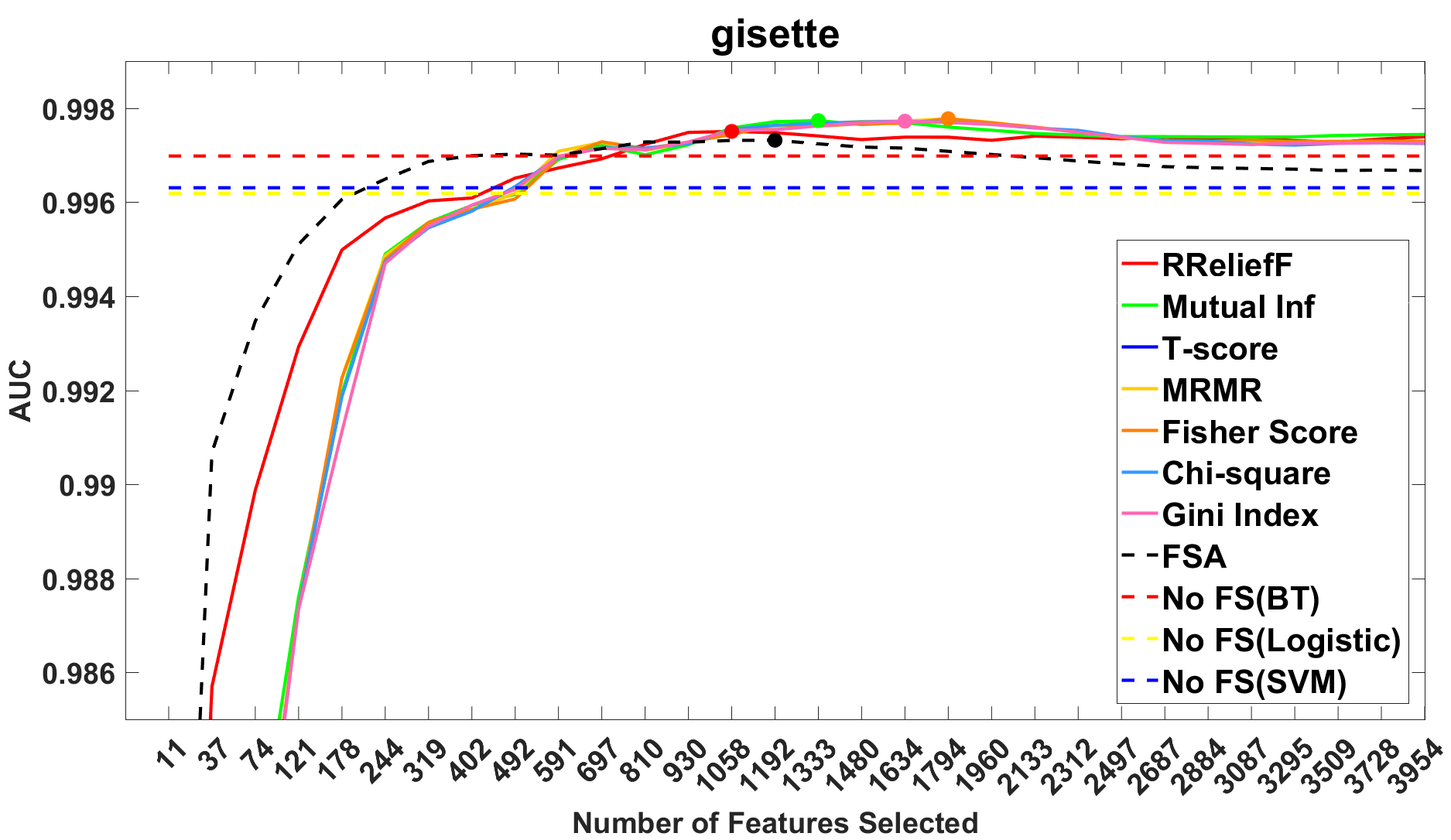}
\includegraphics[width=0.5\linewidth]{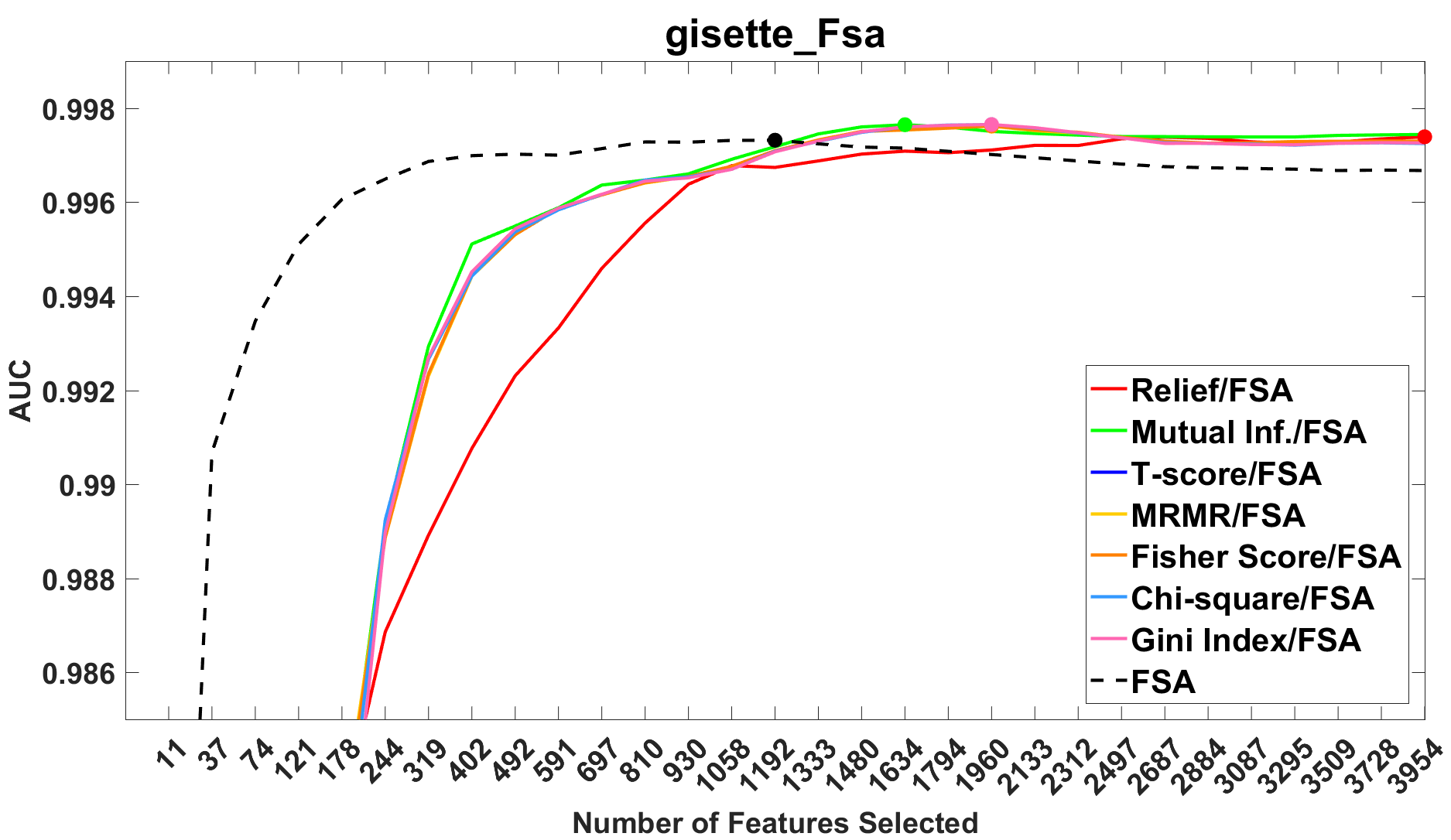}\\
\vspace {-2.5mm}
\hspace{-4mm}
\includegraphics[width=0.5\linewidth]{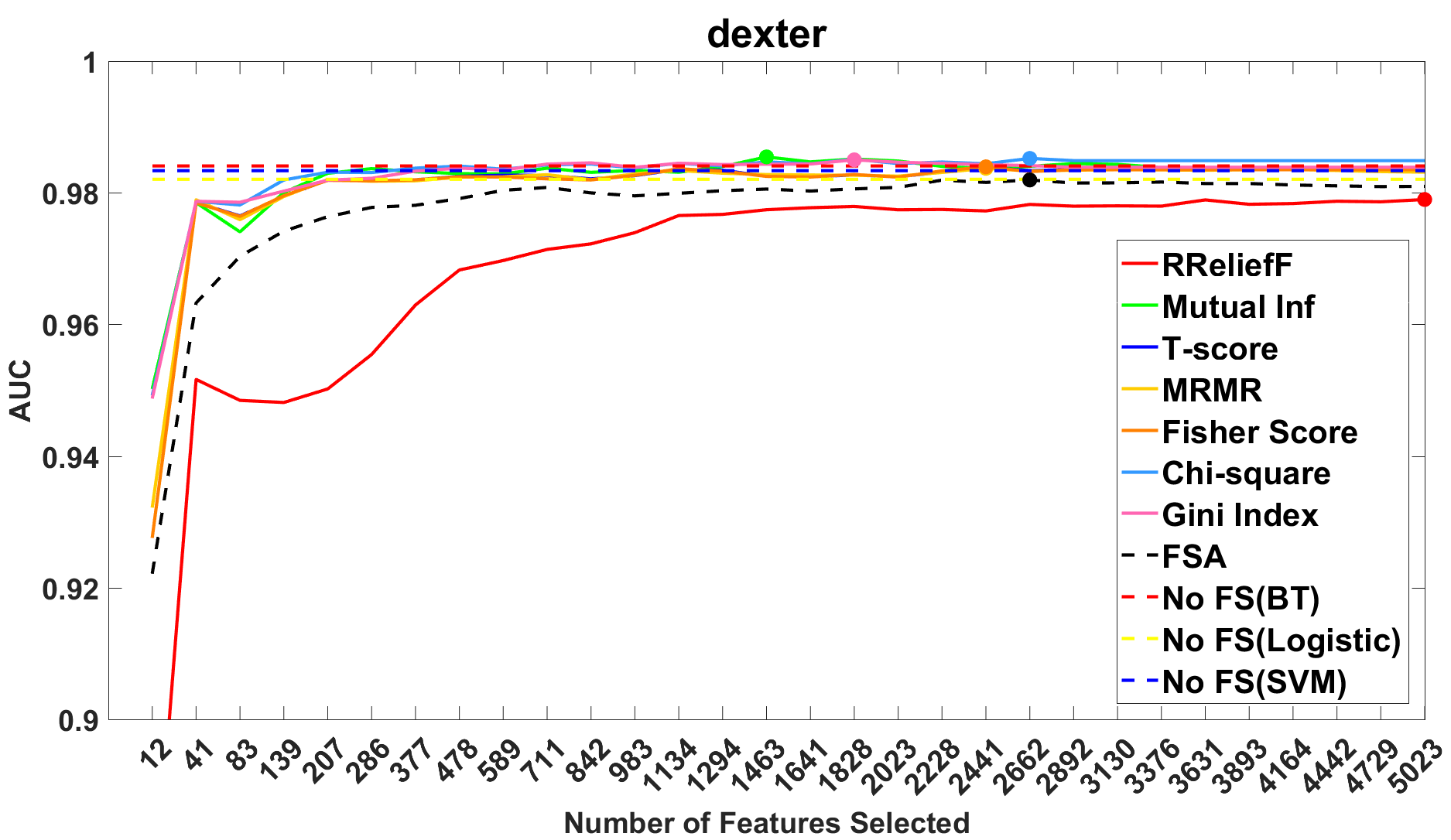}
\includegraphics[width=0.5\linewidth]{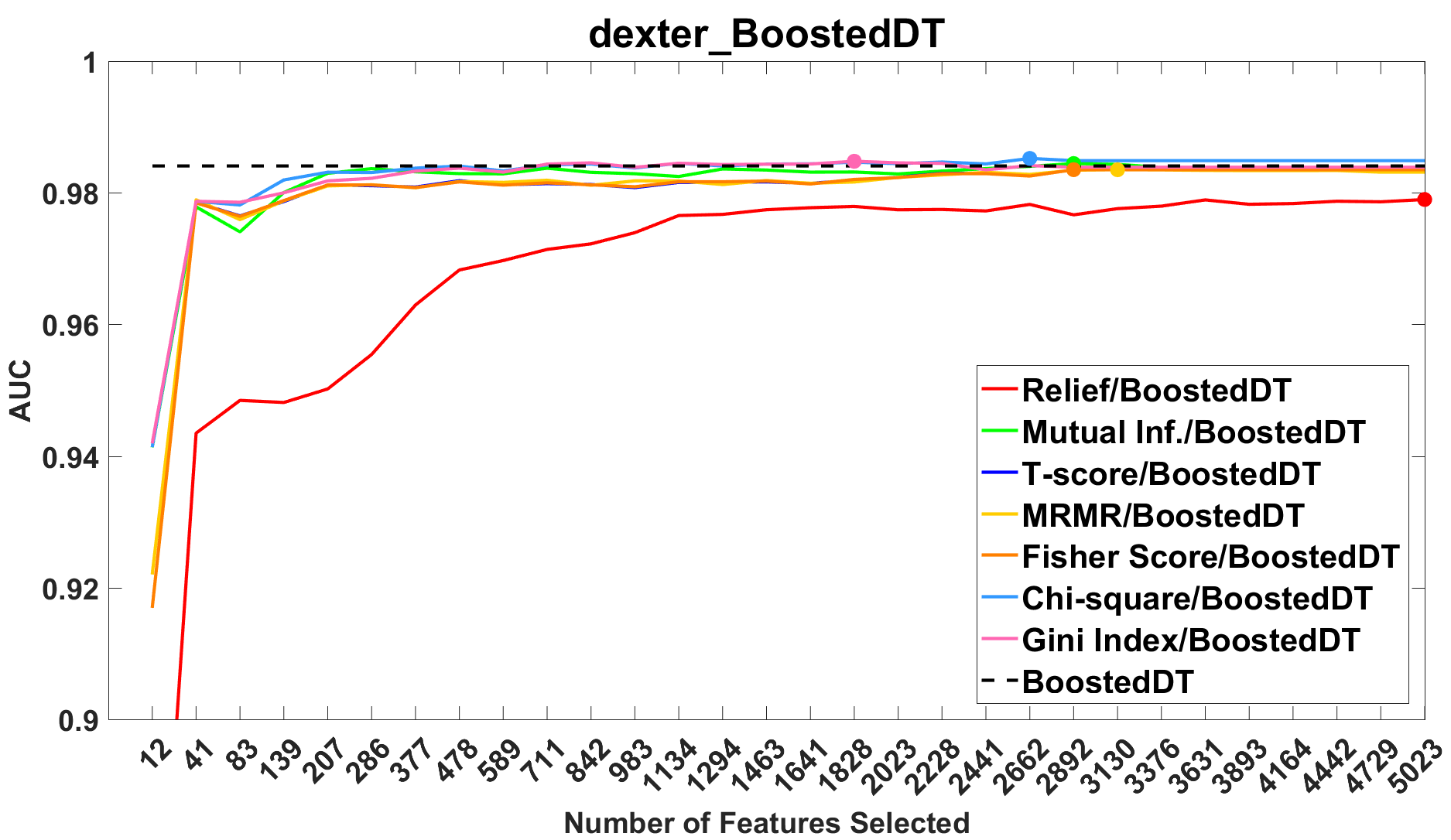}
\caption{Performance plots of methods with and without feature screening. Left: for each screening method are shown the maximum $R^2$ value across all learners. Right: $R^2$ of the screening methods with the best learner for this data.}\label{fig:gisette}
\vspace {-6mm}
\end{figure}

\subsection*{Classification Results}

The following results are based on the output generated by Matlab 2016b. For the methods Relief, T-score, chi-square score, logistic regression, naive Bayes, SVM, boosted decision trees we used their Matlab 2016b implementation. 
For MRMR, Fisher score, and Gini index we used the ASU repository implementation\footnote{\url{http://featureselection.asu.edu/old/software.php}}. Mutual information for classification was implemented by ourselves.
Some of the implementations only accept discrete predictors, so the quantile-based discretization method \cite{discretization} was used.

\subsubsection*{Performance Plots}

In Fig \ref{fig:gisette}, left are shown the AUC of the best learning algorithm vs. the number of features selected by a screening method for four of the classification datasets. 
In Fig \ref{fig:gisette}, right, are shown the AUC of the best overall learning algorithm in each case (SVM for SMK\_CAN\_187, Boosted trees for Madelon and Dexter, FSA for Gisette) vs. the number of features selected by a screening method. 

The plots show that all screening methods help obtain better results on the Gisette and Madelon datasets and most screening methods help obtain better results on the SMK\_CAN\_187 data. 
It can be observed that on the SMK\_CAN\_187 and Madelon, although some screening methods show better results, they select a higher number of selected features than FSA when they reach their optimal values. The right side figure of the SMK\_CAN\_187 plots shows that Relief doesn't work well with the best learner for this dataset. Only three out of the seven methods help obtain a better result on the Dexter dataset.
\begin{figure}[t]
\vspace {-4mm}
\centering
\hspace{-4mm}
\includegraphics[width=0.5\linewidth]{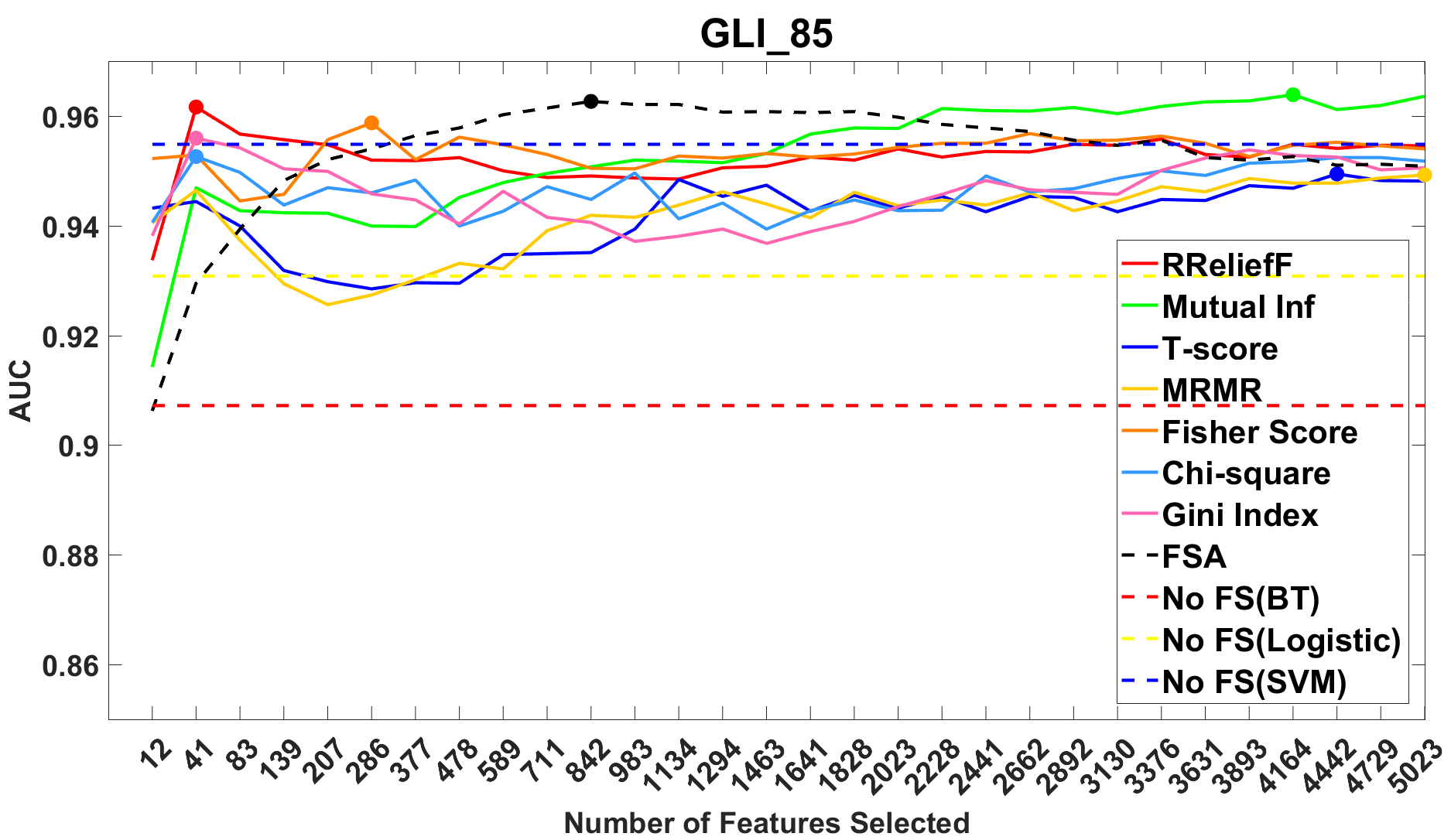}
\includegraphics[width=0.5\linewidth]{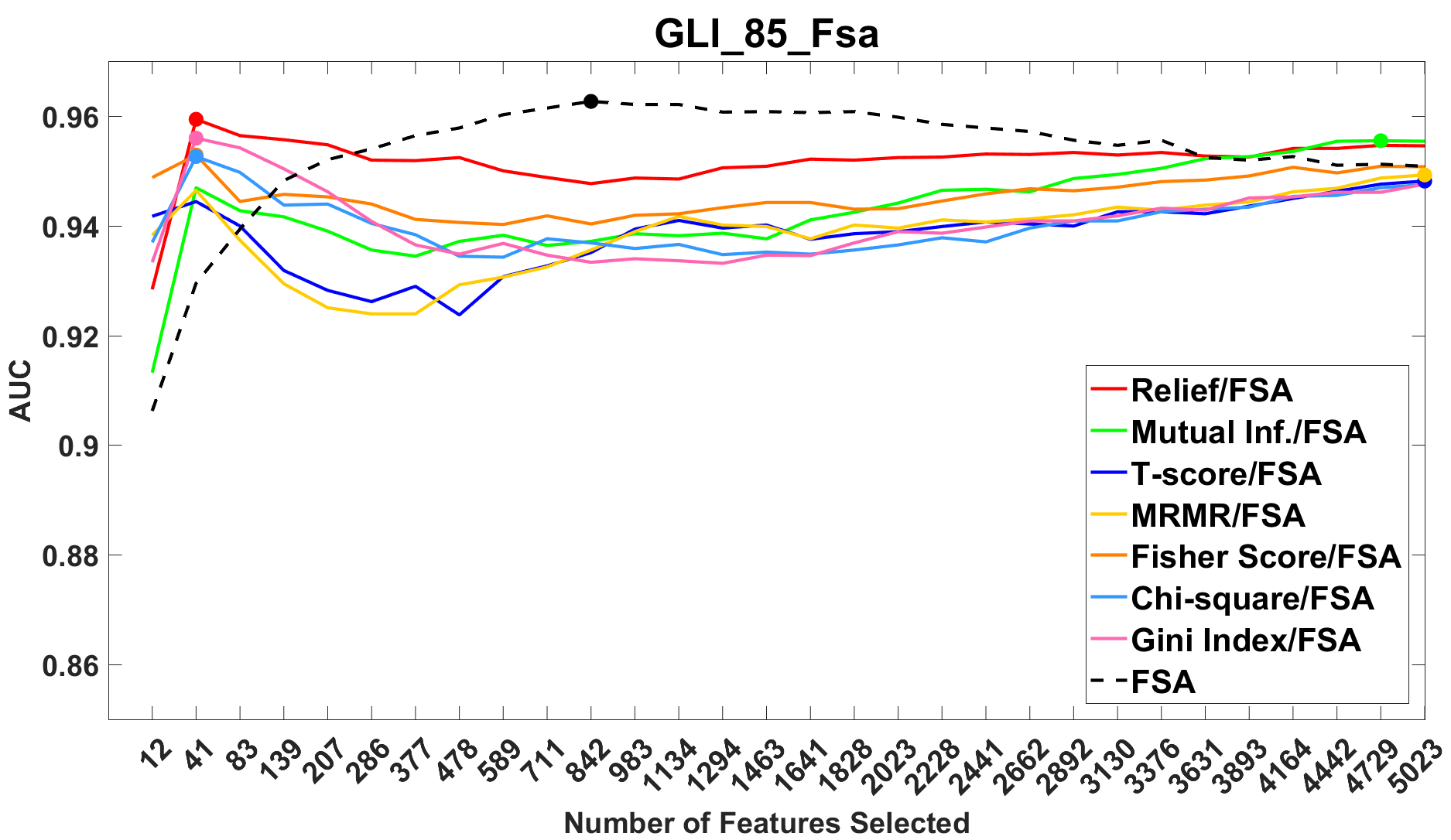}
\caption{Performance plots of methods with and without feature screening. Left: for each screening method are shown the maximum $R^2$ value across all learners. Right: $R^2$ of the screening methods with the best learner for this data.}\label{fig:GLI}
\vspace {-2mm}
\end{figure}

In Fig \ref{fig:GLI}, left are shown the AUC of the best learning algorithm vs. the number of features selected by a screening method on the  GLI\_85 dataset. 
In Fig \ref{fig:GLI}, right, are shown the AUC of FSA (for GLI\_85 ) vs. the number of features selected by a screening method. 
We can observe that only one screening method (mutual information) helps obtain better results than the best learning algorithm without screening.

\begin{table}[htb]
\vspace {-1mm}
\begin{center}
\caption{Overview of the number of datasets where each feature screening method performed significantly better than no screening for different learning algorithms and than the best performing algorithm (larger numbers are better). }
\label{tab:clFilterB}
\begin{tabular}{|l|c|c|c|c|c|c|}
\hline
{Screening Method}& Boost Tree  &FSA &SVM &NB&Logistic &Best algorithm\\
\hline
\hline
{Mutual Information} & 2&1& 3&5&5&2\\
\hline
{Fisher Score \cite{fisherduda2012pattern}} & 3&1&2& 5& 5&2\\
\hline
{Chi-square Score \cite{liu1995chi2}}& 2 &1&2&5& 4&2\\
\hline
{Gini Index \cite{gini1912variability}} & 2&1&2&5& 4&2\\
\hline
{Relief \cite{reliefkira1992feature}} & 3&2&1&5&4&1\\
\hline
{T-score \cite{ttestdavis1986statistics}}& 2&1&2&5&5&2\\
\hline
{MRMR \cite{mrmrhan2005minimum}} & 2&1& 2&5& 4& 2\\

\hline
\end{tabular}
\end{center}
\vspace{-5mm}
\end{table}
\subsubsection*{Comparison Tables}
In Table \ref{tab:clFilterB} is shown the number of datasets on which a filter method helps an algorithm perform significantly better, and the number of datasets on which the filter method helps the best performing learning algorithm perform even better.  
We see that for each learner there is at least one dataset on which a screening method can improve the performance. 
Mutual information, Relief and Fisher score have best performance among all methods. 
It is also clear that the screening methods can generally improve the performance of logistic regression and Naive Bayes  on 4 to 5 out of the 5 datasets. When compared to best leaner without screening methods, Relief shows to be slightly weaker than the others.

\begin{table}[t]
\vspace {-1mm}
\begin{center}
\caption{Ranking of feature screening methods for classification by number of datasets where screening method was significantly better than the best performed no screening method. (larger numbers are better. * indicates appearance.)}
\label{tab:clFilterT}
\begin{tabular}{|l|c|c|c|c|c|c|}
\hline
{Screening Method} &Dexter&Gisette&$SMK\_CAN\_187$&Madelon&$GLI\_85$&Total count \\
\hline
\hline
{Chi-square Score}&=&*&=&*&&2\\
\hline
{Gini Index} &=&*&=&*&=&2\\
\hline
{Relief}&&=&=&*&=&1\\
\hline
{Mutual Information}&=&*&=&*&=&2\\
\hline
{T-score}&=&*&=&*& &2\\
\hline
{Fisher Score}&=&*&=&*&=&2\\
\hline
{MRMR}&=&*&=&*& &2\\
\hline
\end{tabular}
\end{center}
\vspace{-5mm}
\end{table}

In Table \ref{tab:clFilterT} is shown a ``*'' for each dataset and each screening method if it has a learning algorithm that obtains significantly better performance than the best learning algorithm without screening. 
An ``='' sign shows for each dataset whether a screening method is in the same performance group as the best learning algorithm without screening. 
We observe that Relief only worked on the Madelon dataset. The other screening methods worked on both Gisette and Madelon datasets. 
It is also shown that only in a few occasions the screening methods harm the best learning algorithm. 
This is shown by the blank cells in the table. 
Overall, except Relief, the screening methods have similar performance on the five classification datasets. 

\begin{table}[htb]
\vspace {-3mm}
\begin{center}
\caption{Number of datasets where each combination was in the top performing group.}
\label{tab:clA}
\begin{tabular}{|l|c|c|c|c|c|}
\hline
\backslashbox{Filter}{Learners}&Boost Tree& FSA& SVM &NB&Logistic\\
\hline
{Mutual Information} &2&2&2&0&1\\
\hline
{Gini Index} &2&2&1&0&1\\
\hline
{Chi-square Score} &2&2&0&0&1\\
\hline
{Relief} &2&2&0&0&1\\
\hline
{T-score} &1&1&1&0&0\\
\hline
{MRMR} &1&1&1&0&0\\
\hline
{Fisher Score} &1&1&1&0&0\\
\hline
{---} &1&1&1&0&0\\
\hline
\end{tabular}
\end{center}
\vspace {-5mm}
\end{table}

\begin{table}[htb]
\begin{center}
\caption{Ranking of feature screening methods for classification by the number of times each was in the top performing group. (larger numbers are better)}
\label{tab:ClFilterA}
\vskip 1mm
\begin{tabular}{|l|c|c|}
\hline
 &\multicolumn{2}{c|}{Top performing}\\ 
{Screening Method} &Method-Algorithm &Method\\
\hline
\hline
{Mutual Information} &7&4\\
\hline
{Gini Index} &6&4\\
\hline
{Chi-square Score} &5&3\\
\hline
{Relief} &5&3\\
\hline
{T-score} &3&2\\
\hline
{MRMR} &3&2\\
\hline
{Fisher Score} &3&2\\
\hline
{No Screening} &3&3\\
\hline
\end{tabular}
\end{center}
\vspace{-5mm}
\end{table}
In Table \ref{tab:clA} is shown for each screening method-learning algorithm combination the number of datasets for which it was in the top performing group.
We can observe that the screening methods with boosted trees and FSA have the overall best performance. Among them, boosted trees and FSA with four screening methods (Chi-square Score, Gini Index, Relief and Mutual Information) have a slight advantage compared to the algorithms without screening. SVM worked well with Mutual Information. The above named four screening methods also helped Logistic regression on one dataset. Naive Bayes didn't perform well on these five datasets.

In Table \ref{tab:ClFilterA}, are shown  the number of times each screening method was in the top performing group. In the first column, these methods were counted with respect to the learning algorithms they were applied. So there can be at most 25 counts (for each screening method there are five learning algorithms and five datasets) in each cell. The second column shows the counts with the best learning algorithm, so there can be at most 5 counts in each cell. The Mutual Information has the highest counts. It's significantly higher than no screening. Gini Index, Relief and Chi-square score also have relative higher counts when considering them together with a  learning algorithm. Mutual Information and Gini Index have good performance on more datasets than using no screening, when considering only the best learning algorithm for each method and each dataset.

\section*{Discussion}

Since we are interested in evaluating screening methods on real datasets, we don't have information about the true features that are relevant in connection with the response, so we can only look at prediction performance. In this respect, there are at least two ways to see whether the screening methods are useful for real datasets. 

If we ask whether they are helpful in improving the prediction performance of the best learning algorithm from our arsenal, then the answer is ``Some of them are sometimes useful, on two datasets out of five, in both regression and classification''. 
Indeed, for regression we see from Table \ref{tab:regFilterT} that Mutual Information and RReliefF were helpful in improving the prediction of the best learning algorithm on two datasets out of five, while the other two screening methods were only helpful on one dataset. 
For classification, we see from Table \ref{tab:clFilterT} that most screening methods were helpful in improving the prediction of the best learning algorithm on two datasets out of five, except Relief, which was only helpful on one dataset.

If however we are interested in using a screening method to reduce the dataset size, then we might ask whether we lose any prediction performance this way. In this case our answer would be ``Usually not, for the right screening method, especially in classification''. In Table \ref{tab:regFilterT} and Table \ref{tab:clFilterT} are shown that only in very few occasions do the screening methods harm the performance of best learning algorithm. For regression, we see from Table \ref{tab:regFilterA} that RReliefF is the best in this respect, remaining in the top performing group (with the right algorithm and number of selected features) on 4 out of 5 regression datasets. For classification, from Table \ref{tab:ClFilterA}  we see that Mutual Information and Gini index are the best, remaining in the top performing group (with the right algorithm and number of selected features) on 4 out of 5 classification datasets.

If we had to select one screening method that is most successful at both of these tasks, this method would be Mutual Information. We see that it is the only method that is helpful in improving performance in both regression and classification, and stays in the top performing group on most datasets, for both regression and classification.

\section*{Conclusion}
Some of the screening methods that were evaluated in this paper bring an improvement in prediction for some datasets, in both regresison and classification.  In the classification tasks, the screening methods with boosted trees give the best overall results. 
All the seven classification screening methods evaluated help improve the performance of learner to a certain degree.
The Mutual Information , Gini Index, Chi-square score and Relief work slightly better than the other methods. 
It also can be seen from the tables that the screening methods work well especially on learning algorithms that give poor results on their own. 
Compared to classification, there are fewer screening methods for regression problems. Of the three regression screening methods evaluated, RReliefF and Mutual Information work better than correlation, and improve the best learning algorithm performance on two datasets out of five.

\section*{Acknowledgments}
The authors declare that there are no conflicts of interest.

\nolinenumbers

%
%
%

\bibliography{references}

\begin{thebibliography}{10}

\bibitem{ttestdavis1986statistics}
Davis JC, Sampson RJ.
\newblock Statistics and data analysis in geology. vol. 646.
\newblock Wiley New York et al.; 1986.

\bibitem{mutuallewis1992feature}
Lewis DD.
\newblock Feature selection and feature extraction for text categorization.
\newblock In: Proceedings of the workshop on Speech and Natural Language.
  Association for Computational Linguistics; 1992. p. 212--217.

\bibitem{reliefkira1992feature}
Kira K, Rendell LA.
\newblock The feature selection problem: Traditional methods and a new
  algorithm.
\newblock In: AAAI. vol.~2; 1992. p. 129--134.

\bibitem{tibshirani1996regression}
Tibshirani R.
\newblock Regression shrinkage and selection via the lasso.
\newblock Journal of the Royal Statistical Society Series B (Methodological).
  1996; p. 267--288.

\bibitem{mrmrhan2005minimum}
Han C, Chris D, Fu H.
\newblock Minimum redundancy maximum relevance feature selection [J].
\newblock IEEE Intelligent Systems. 2005;20(6):70--71.

\bibitem{li2016feature}
Li J, Cheng K, Wang S, Morstatter F, Trevino RP, Tang J, et~al.
\newblock Feature selection: A data perspective.
\newblock ACM Computing Surveys (CSUR). 2017;50(6):94.

\bibitem{tang2014feature}
Tang J, Alelyani S, Liu H.
\newblock Feature selection for classification: A review.
\newblock Data Classification: Algorithms and Applications. 2014; p.~37.

\bibitem{chandrashekar2014survey}
Chandrashekar G, Sahin F.
\newblock A survey on feature selection methods.
\newblock Computers \& Electrical Engineering. 2014;40(1):16--28.

\bibitem{jovic2015review}
Jovi{\'c} A, Brki{\'c} K, Bogunovi{\'c} N.
\newblock A review of feature selection methods with applications.
\newblock In: 2015 38th International Convention on Information and
  Communication Technology, Electronics and Microelectronics (MIPRO). IEEE;
  2015. p. 1200--1205.

\bibitem{cai2018feature}
Cai J, Luo J, Wang S, Yang S.
\newblock Feature selection in machine learning: A new perspective.
\newblock Neurocomputing. 2018;300:70--79.

\bibitem{li2017recent}
Li Y, Li T, Liu H.
\newblock Recent advances in feature selection and its applications.
\newblock Knowledge and Information Systems. 2017;53(3):551--577.

\bibitem{urbanowicz2018benchmarking}
Urbanowicz RJ, Olson RS, Schmitt P, Meeker M, Moore JH.
\newblock Benchmarking relief-based feature selection methods for
  bioinformatics data mining.
\newblock Journal of biomedical informatics. 2018;85:168--188.

\bibitem{alelyani2013feature}
Alelyani S, Tang J, Liu H.
\newblock Feature Selection for Clustering: A Review.
\newblock Data Clustering: Algorithms and Applications. 2013;29:110--121.

\bibitem{talavera2005evaluation}
Talavera L.
\newblock An evaluation of filter and wrapper methods for feature selection in
  categorical clustering.
\newblock Advances in Intelligent Data Analysis VI. 2005; p. 742--742.

\bibitem{masoudi2019featureselect}
Masoudi-Sobhanzadeh Y, Motieghader H, Masoudi-Nejad A.
\newblock FeatureSelect: a software for feature selection based on machine
  learning approaches.
\newblock BMC bioinformatics. 2019;20(1):170.

\bibitem{chen2018ifeature}
Chen Z, Zhao P, Li F, Leier A, Marquez-Lago TT, Wang Y, et~al.
\newblock iFeature: a python package and web server for features extraction and
  selection from protein and peptide sequences.
\newblock Bioinformatics. 2018;34(14):2499--2502.

\bibitem{guyon2003introduction}
Guyon I, Elisseeff A.
\newblock An introduction to variable and feature selection.
\newblock Journal of machine learning research. 2003;3(Mar):1157--1182.

\bibitem{sanchez2007filter}
S{\'a}nchez-Maro{\~n}o N, Alonso-Betanzos A, Tombilla-Sanrom{\'a}n M.
\newblock Filter methods for feature selection--a comparative study.
\newblock Intelligent Data Engineering and Automated Learning-IDEAL 2007. 2007;
  p. 178--187.

\bibitem{saeys2007review}
Saeys Y, Inza I, Larra{\~n}aga P.
\newblock A review of feature selection techniques in bioinformatics.
\newblock bioinformatics. 2007;23(19):2507--2517.

\bibitem{rrelieffrobnik1997adaptation}
Robnik-{\v{S}}ikonja M, Kononenko I.
\newblock An adaptation of Relief for attribute estimation in regression.
\newblock In: Machine Learning: Proceedings of the Fourteenth International
  Conference (ICML’97); 1997. p. 296--304.

\bibitem{barbu2017FSA}
Barbu A, She Y, Ding L, Gramajo G.
\newblock Feature Selection with Annealing for Computer Vision and Big Data
  Learning.
\newblock IEEE Transactions on Pattern Analysis and Machine Intelligence.
  2017;39(2):272--286.

\bibitem{liu1995chi2}
Liu H, Setiono R.
\newblock Chi2: Feature selection and discretization of numeric attributes.
\newblock In: Tools with artificial intelligence, 1995. proceedings., seventh
  international conference on. IEEE; 1995. p. 388--391.

\bibitem{fisherduda2012pattern}
Duda RO, Hart PE, Stork DG.
\newblock Pattern classification.
\newblock John Wiley \& Sons; 2012.

\bibitem{gini1912variability}
Gini C.
\newblock Variability and mutability, contribution to the study of statistical
  distribution and relaitons.
\newblock Studi Economico-Giuricici della R. 1912;.

\bibitem{wang2006genetic}
Wang S, Yehya N, Schadt EE, Wang H, Drake TA, Lusis AJ.
\newblock Genetic and genomic analysis of a fat mass trait with complex
  inheritance reveals marked sex specificity.
\newblock PLoS genetics. 2006;2(2):e15.

\bibitem{tumorNCIGDC}
Grossman RL, Heath AP, Ferretti V, Varmus HE, Lowy DR, Kibbe WA, et~al.
\newblock Toward a Shared Vision for Cancer Genomic Data.
\newblock New England Journal of Medicine. 2016;375(12):1109--1112.

\bibitem{torres2014ujiindoorloc}
Torres-Sospedra J, Montoliu R, Mart{\'\i}nez-Us{\'o} A, Avariento JP, Arnau TJ,
  Benedito-Bordonau M, et~al.
\newblock UJIIndoorLoc: A new multi-building and multi-floor database for WLAN
  fingerprint-based indoor localization problems.
\newblock In: Indoor Positioning and Indoor Navigation (IPIN), 2014
  International Conference on. IEEE; 2014. p. 261--270.

\bibitem{Rothe-IJCV-2016}
Rothe R, Timofte R, Gool LV.
\newblock Deep expectation of real and apparent age from a single image without
  facial landmarks.
\newblock International Journal of Computer Vision (IJCV). 2016;.

\bibitem{CoEPrA20063}
Ivanciuc O. {CoEPrA 2006 Round 3} Comparative Evaluation of Prediction
  Algorithms; 2006.
\newblock Available from: \url{http://www.coepra.org/}.

\bibitem{guyon2005result}
Guyon I, Gunn S, Ben-Hur A, Dror G.
\newblock Result analysis of the NIPS 2003 feature selection challenge.
\newblock In: Advances in neural information processing systems; 2005. p.
  545--552.

\bibitem{spira2007airway}
Spira A, Beane JE, Shah V, Steiling K, Liu G, Schembri F, et~al.
\newblock Airway epithelial gene expression in the diagnostic evaluation of
  smokers with suspect lung cancer.
\newblock Nature medicine. 2007;13(3):361.

\bibitem{freije2004gene}
Freije WA, Castro-Vargas FE, Fang Z, Horvath S, Cloughesy T, Liau LM, et~al.
\newblock Gene expression profiling of gliomas strongly predicts survival.
\newblock Cancer research. 2004;64(18):6503--6510.

\bibitem{uciML}
Lichman M. {UCI} Machine Learning Repository; 2013.
\newblock Available from: \url{http://archive.ics.uci.edu/ml}.

\bibitem{parkhi2015deep}
Parkhi OM, Vedaldi A, Zisserman A, et~al.
\newblock Deep Face Recognition.
\newblock In: BMVC. vol.~1; 2015. p.~6.

\bibitem{Matlab2016b}
MATLAB Release 2016b; 2016.

\bibitem{discretization}
Nguyen(2014) XV. Information Theoretic Feature Selection, version 1.1; Updated
  07 Jul 2014.
\newblock Available from:
  \url{https://www.mathworks.com/matlabcentral/fileexchange/47129-information-theoretic-feature-selection}.

\end{thebibliography}

\section*{Supporting information}
\subsection*{Review of Screening Methods}\label{sec2:rv}
In this section we give an overview of some of the existing screening (filter) methods for classification and regression, which will be evaluated in our experiments.

\subsubsection*{List of Symbols}
The following short list of symbols are used throughout the document.
\begin{center}
\begin{tabular}{ll}
$n$& the number of observations\\
$p$& the number of variables\\
$k$& the number of true features\\
$S=\{(\bx_i,y_i)\in \RR^p\times \RR,i=1,...,n\}$& the data space\\
$X$&  the $n\times p$ data matrix\\
$X_j, j=1,...,p$& the $j$-th column/feature of $X$\\
$\bx_i, i=1,...,n$& the i-$th$ observation of $X$\\
$x_{ij}, i=1,...,n j=1,...,p$& the $j$-th column/feature of the i-$th$ observation of $X$\\
$\by$& the $n\times 1$ target vector\\
$y_i, i=1,...,n$& the $j$-th target value\\
\end{tabular}
\end{center}

\setcounter{tocdepth}{4}
\setcounter{secnumdepth}{4}
\subsubsection*{Screening Methods for Classification}
\paragraph*{Mutual Information} \label{sec:clmutual}
The mutual information (a.k.a. information gain) method measures the information shared by two variables of interest, in this case, a feature $X_j$ and the class label $\by$. The mutual information between variable $A$, where $S_A=\{A \in \RR\}$  and variable $Y$, where $S_Y=\{Y \in \RR\}$ can be described as:
\vspace{5mm}
\begin{equation}
I(A,Y)=\int_{S_A} \int_{S_Y} p(A,Y)\log\frac{p(A,Y)}{p(A)p(Y)}dA dY\label{eq:mutual_popul}
\end{equation}
where $p(A,Y)$ is the joint probability density of $A$ and $Y$, while $p(A)$ and $p(Y)$ are the marginal p.d.f.s of $A$ and $Y$.

In practice, given a sample dataset, each feature can be discretized into bins based on the value range. Here, $b=1,2,...,B$ indicates bin number, $c=1,2,...,C$ indicates class number. Therefore mutual information between label vector $\by$ and feature vector $X_j$ can also be described as:
\vspace{5mm}
\begin{equation}
I(X_j,\by)=\sum_{b=1}^B \sum_{c=1}^C p({X_j}_b,\by_c)\log\frac{p({X_j}_b,\by_c)}{p({X_j}_b)p(\by_c)}
\label{eq:mutual_sample}
\end{equation}
where $p({X_j}_b,\by_c)$ is the joint probability of bin ${X_j}_b$ and label vector $\by_c$, while $p({X_j}_b)$ and $p(\by_c)$ are the marginal probabilities. Features that are more related to the classification label tend to have higher mutual information.

\paragraph*{Relief  and ReliefF} \label{sec:relief}
The idea of the Relief algorithm is to measure how well a feature's values can distinguish instances that are near each other. For the $i$-th instance-label pair $(\bx_i,y_i)$, denote its nearest instance neighbor from the same class as the nearest hit $(\bx^{hit}_i,y_i)$, and its nearest instance neighbor from a different class as the nearest miss $(\bx_i^{miss},y_i^{miss})$. The distance between two instances $\bx_i,\bx_j$ is calculated using the Euclidean norm $\|\bx_i-\bx_j\|$.  Then the Relief measure for a certain feature $F$ can be computed as:
\vspace{5mm}
where the function $\diff(F: x,y)$ calculates the difference between the values of feature $F$ for two instances. For discrete features $\diff(F: x,y)$ is defined as:
\vspace{5mm}
\begin{equation}
    \diff(F:x,y)=
    \begin{cases}
      0;  & \text{if }x=y \\
      1; & \text{otherwise}
    \end{cases}\label{eq:diff1}
\end{equation}
and for a continuous feature $X_j$ as:
\vspace{5mm}
\begin{equation}
\diff(F:x,y)=\frac{|x-y|}{\max(F)-\min(F)}\label{eq:diff2}
\end{equation}
The Relief measure can also be extended to a multi-class version ReliefF, but we are only interested in binary classification in this paper.
In summary, higher Relief values indicate better discrimination power of the label by the feature values.

\paragraph*{Minimum Redundancy Maximum Relevance} \label{sec:mrmr}
The minimum redundancy maximum relevance (MRMR) method is set to choose the feature that has the highest mutual information difference (MID) or mutual information quotient (MIQ). The MID and MIQ are calculated as :
\vspace{5mm}
\begin{equation}
MID_j=I(X_j,\by)-\frac{1}{|Q|}\sum_{q\in Q}I(X_j,X_q)\label{eq:mrmrmid}
\end{equation}
\begin{equation}
MIQ_j=\frac{I(X_j,\by)}{\frac{1}{|Q|}\sum_{q\in Q}I(X_j,X_q)}\label{eq:mrmrmiq}
\end{equation}
where $Q$ is the set of features already selected, $I(X_j,\by)$ is the mutual information for $j$-th feature and the label vector $\by$, and $I(X_j,X_q)$ denotes the mutual information between features $j$ and $q$. 

In the case where the features take continuous values, MIQ and MID can be modified as the F-test correlation difference (FCD) and F-test correlation quotient (FCQ). FCD and FCQ are computed as:
\vspace{5mm}
\begin{equation}
FCD_j=F(X_j,\by)-\frac{1}{|Q|}\sum_{q\in Q}|c(X_j,X_q)|\label{eq:mrmrfcd}
\end{equation}
\vspace{5mm}
\begin{equation}
FCQ_j=\frac{F(X_j,\by)}{\frac{1}{|Q|}\sum_{q\in Q}|c(X_j,X_q)|}\label{eq:mrmrfcq}
\end{equation}
where $F(X_j,\by)$ is the F-statistic for $j$-th feature and label vector $\by$, and $|c(X_j,X_q)|$ denotes the absolute correlation coefficient between features $j$ and $q$. In the case of binary labels the F-statistic can be replaced by the T-statistic.

\paragraph*{T-Score} \label{sec:tscore}
The T-score method is a feature screening method applied on datasets with binary labels. The method is based on the calculation of the $t$-statistic. The basic idea is to divide each feature's values into two sample groups based on their labels. Then the $t$-statistic is calculated to examine if the two sample groups have statistically significant differences in their means.
For each feature $X_j$, the values of $X_j$ are divided into two groups based on their labels. Then the means $\mu_1$ and $\mu_2$ are calculated as the means of the two groups and $\sigma_1$ and $\sigma_2$ are standard deviations of these two groups respectively. Let $n_1$ and $n_2$ be the number of instances of the two groups. Then the $t$-statistic for feature $i$ can be calculated as:
\vspace{5mm}
\begin{equation}
T_j=\frac{|\mu_1-\mu_2|}{\sqrt{\frac{\sigma_1^2}{n_1}+\frac{\sigma_2^2}{n_2}}}\label{eq:T}
\end{equation}
Generally speaking, the higher the $t$-statistic, the more separated the two labels are by values of that feature and therefore the more relevant that feature is for classification.

\paragraph*{Chi-square Score} \label{sec:chi2}
The chi-square score method is based on the chi-square statistic. It can test the independence between two variables, therefore it can also test the relevance of a variable $X_j$ for the label vector $\by$. If feature $X_j$ has $L$ levels (discretized if necessary) and $\by$ has $C=2$ levels (label categories),  let $n_{lc}$ denote the number of instances with label $c$ and level $l$ for feature $j$. Let $\hat{n}_{lc}$ denote the estimated number of instances with label $c$ and having level $l$, $\hat{n}_{lc}=\frac{n_{l}n_{c}}{n}$, where $n$ is the total number of instances, $n_{l}$ is the number of instances having level $l$, and $n_{c}$ is the number of instances with label $c$. The chi-square statistic is then computed as:
\begin{equation}
\chi^2_j=\sum_{l=1}^L\sum_{c=1}^C\frac{(n_{lc}-\hat{n}_{lc})^2}{\hat{n}_{lc}}\label{eq:chi2}
\end{equation}
Usually, a higher chi-square statistic indicates low independence, in other word, a higher relevance between that feature and label.

\paragraph*{Gini Index} \label{sec:gini}
The Gini index method is based on the Gini impurity after splitting a sample set. For a given feature $X_j$, let 
${A}_{h}=\{i, x_{ij}\leq h\}$ denote the instances whose values of the $j-$th feature is smaller than or equal to $h$ and ${B}_{h}=\{i, x_{ij}> h\}$.
The Gini impurity for subset ${A}_{h}$ or ${B}_{h}$ can be expressed as:
\vspace{5mm}
\begin{equation}
Gini(A_h)=1-\sum_{c=1}^CP(C_c|A_h)^2\label{eq:gini1}
\end{equation}
where $C$ is the number of labels and  $c \in \{1,2,...,C\}$ are the  label categories. $P({C}_{c}|{A}_{h})$ is the conditional probability of instances having label $c$ given that they are in subset  ${A}_{h}$. Let ${a}_{c}$ denote the number of instances in ${A}_{h}$ with label $c$. Let ${a}_{h}$ denote the number of instances in ${A}_{h}$. Then $P({C}_{c}|{A}_{h})$ can be calculated as ${a}_{c}$/${a}_{h}$.

Based on these notations, the Gini index after splitting is:
\vspace{5mm}
\begin{equation}
Gini_{split}=P(A_h)Gini(A_h)+P(B_h)Gini(B_h)\label{eq:gini2}
\end{equation}
where P(${A}_{h}$) is the number of instances in subset ${A}_{h}$ divided by the number of total instances. 
Therefore for each feature, the Gini index can be calculated as:
\vspace{5mm}
\begin{equation}
Gini_j=P(A_h)(1-\sum_{c=1}^CP(C_c|A_h)^2)+P(B_h)(1-\sum_{c=1}^CP(C_c|B_h)^2)\label{eq:gini3}
\end{equation}
Basically, the Gini index measures the frequency that a randomly chosen instance from the sample set would be incorrectly labeled. 
So for all possible thresholds $h$ of one feature, select the minimum Gini index as this feature's Gini index. Features with smaller Gini index are preferred.

\paragraph*{Fisher Score} \label{sec:fisher}
The idea of the Fisher score is to choose the feature subset, for which the observations have the largest possible between class distances and the smallest possible within class distances. This would be the feature subset that has the largest Fisher score. The Fisher score for any feature set is computed as:
\vspace{5mm}
\begin{equation}
Fisher=Tr(D_b)(D_t+\gamma I)^{-1}\label{eq:fisherall}
\end{equation}
where $\gamma$ is a regularization term, $D_b$ is called between-class scatter matrix, $D_t$ is called total scatter matrix. Since for a certain feature subset with size $d$, there are ${m \choose d}$ combinations of Fisher scores to be calculated, this is too computationally expensive. 
For this reason, a heuristic is to compute the scores for each feature with respect to the Fisher score criterion.
The individual Fisher score is computed as:
\vspace{5mm}
\begin{equation}
Fisher_j=\frac{\sum_{c=1}^Cn_c(\mu_c-\mu)^2}{\sum_{c=1}^Cn_c\sigma_c^2}\label{eq:fisherindividual}
\end{equation}
where $\mu$ and $\sigma$ are mean and standard deviation of that feature, and $\mu_c$ is the mean of the feature values for observations with label $c$ and $n_c$ is the number of instances with label $c$.
Features with larger Fisher scores are preferred.

\subsubsection*{Screening Methods for Regression}

\paragraph*{Correlation} \label{sec:corr}
The correlation feature screening method is based on the calculation of correlation coefficient between response and features. It is evaluated as following:
\vspace{5mm}
\begin{equation}
\rho_j=\frac{cov(X_j,\by)}{\sigma_\by\sigma_{X_j}}\label{eq:corr}
\end{equation}
where $X_j$ is $j-$th feature, $\by$ is response.
Features with larger correlation coefficient are preferred.

\paragraph*{Mutual Information} \label{sec:regmutual}
To apply mutual information for regression data, we discretize both the feature and the response into a numbers of bins. For feature $X_j$ and response $\by$, let $x_{jb}$ and $y_l$ indicate values falling in $b$-th and $l$-th bins respectively. 
The mutual information for the $j$-th feature is computed as:
\vspace{5mm}
\begin{equation}
I(X_j,Y)=\sum_{b=1}^B\sum_{l=1}^LP(x_{jb},y_l)\log\frac{P(x_{jb},y_l)}{P(x_{jb})P(y_l)}\label{eq:mutualr}
\end{equation}
Let $n$ denote the number of instances. Then $P(x_{jb},y_l)$ can be estimated by $N_{jbl}$/n, where $N_{jkl}$ is the number of instances falling into feature bin $b$ and response bin $l$. 
Also, $P(x_{jb})$ can be estimated by $N_{jb}$/n, where $N_{jb}$ is the number of instances lay in feature bin $b$, and $P(y_l)$ can be estimated by $N_{l}$/n, where $N_{l}$ is the number of instances lay in response bin $l$. 
Features with larger mutual information have more influence on the response.

\paragraph*{RReliefF} \label{sec:rrelieff}
RReliefF is a regression version of Relief. It starts from the original weight function. For feature $A$ the function can be expressed as:
\begin{equation}
\begin{split}
W(A)=P(\text{different value of A}|\text{nearest instance from different class})\\
         -P(\text{different value of A}|\text{nearest instance from the same class})\label{eq:RreliefF}
\end{split}
\end{equation}
Denote
\begin{equation}
\begin{split}
P_{diffA}&=P(\text{different value of A}|\text{nearest instances})\\
\label{eq:diffa}
P_{diffP}&=P(\text{different response}|\text{nearest instances})\\
P_{diffP|diffA}&=P(\text{different response}|\text{different value of A and nearest instances}).
\end{split}
\end{equation}
Then from \eqref{eq:RreliefF}, using Bayes' rule:
\begin{equation}
W(A)=\frac{P_{diffP|diffA}P_{diffA}}{P_{diffP}}-\frac{(1-P_{diffP|diffA})P_{diffA}}{1-P_{diffP}},
\label{eq:RreliefB}
\end{equation}
which can be further modified as:
\begin{equation}
W(A)=\frac{N_{dP\&dA}}{N_{dP}}-\frac{(N_{dA}-N_{dP\&dA})}{m-N_{dP}}
\label{eq:RreliefM}
\end{equation}
where $N_{dA}$, $N_{dP}$ and $N_{dP\&dA}$ denote different feature value, different response value and different feature \&  response value respectively.
Denote for instance $\bx_i$ its $k$-nearest instances as $\bu_{ij}, j\in\{1,...,k\}$. Then the expressions for $N_{dA}$, $N_{dP}$ and $N_{dP\&dA}$ are:
\begin{equation}
\begin{split}
N_{dA}&=\sum_{i=1}^n \sum_{j=1}^k \diff(A:\bx_i,\bu_{ij})d(i,j)\\
\label{eq:nda}
N_{dP}&=\sum_{i=1}^n \sum_{j=1}^k\diff(y:\bx_i,\bu_{ij})d(i,j)\\
N_{dP\&dA}&=\sum_{i=1}^n \sum_{j=1}^k\diff(y:\bx_i,\bu_{ij})\diff(A:\bx_i,\bu_{ij})d(i,j)\\
\end{split}
\end{equation}
Where $\diff(F,x,y)$ is defined in Eq. \eqref {eq:diff1} and \eqref{eq:diff2} and $d(i,j)$ is used to take account the distance between $\bx_i$ and $\bu_j$:
\begin{equation}
d(i,j)=\frac{d_1(i,j)}{\sum_{l=1}^kd_1(i,l)}
\label{eq:d}
\end{equation}
and 
\begin{equation}
d_1(i,j)=\exp(-\text{rank}^2(\bx_i,\bu_{ij})/\sigma^2)
\label{eq:d1}
\end{equation}
where rank$(\bx_i, \bu_{ij})$ is the rank of the instance $\bu_{ij}$ in a sequence of instances ordered by the distance from $\bx_i$, and $\sigma$ is a user defined parameter. $d_1(i,j)$ is calculated in an exponentially decreasing fashion with the idea that further instances should have lesser influence. Usually, $d_1(i,j)$ takes value $1/k$.
Features with larger $W(\cdot)$ are preferred.

\subsubsection*{Feature Selection With Annealing (FSA)} \label{sec:fsa}
Feature Selection With Annealing (a.k.a. FSA) is a recent embedded method for feature selection that can handle high dimensional data. 
FSA can bring the relevant feature space down to an acceptable level using an variable removal schedule and obtain a rather accurate and stable model. 
The basic algorithm of FSA is:
\begin{algorithm}[htb]
   \caption{{\bf Feature Selection with Annealing (FSA)}}
   \label{alg:fsa}
\begin{algorithmic}
   \STATE {\bfseries Input:} Training samples $(\bx_i,y_i)\hspace{-1mm}\in \hspace{-1mm}\RR^p\times\RR, i\hspace{-0.5mm}=\hspace{-0.5mm}1, 2, ..., N$.
   \STATE {\bfseries Output:} Trained model parameter vector $\bbeta$.
\end{algorithmic}
\begin{algorithmic} [1]
\STATE Initialize $\bbeta$.
        \FOR {e=1 to $N^{iter}$}
                \STATE  Update $\bbeta \leftarrow \bbeta-\eta \frac{\partial L(\bbeta)}{\partial \bbeta}$
                \STATE Keep the $M_e$ features with highest $|\bbeta_j|$ and renumber them 1, ..., $M_e$.
      \ENDFOR
\end{algorithmic}
\end{algorithm}

The value of $N^{iter}$ in step 2 is the total number of iterations. The formula in step 3 uses a typical gradient descent or an epoch of stochastic gradient descent with momentum and minibatch towards minimizing the loss $L(\bbeta)$. 
The $M_e$ in step 4 is the annealing schedule which gradually decreases with the iteration number $e$. It decides how many features to keep in each iteration. 
Let $k$ be a user defined parameter controlling how many features to keep in the end. The $M_e$ can be computed as:
\begin{equation}
M_e=k+(p-k)\max(0,\frac{N^{iter}-2e}{2e\mu+N^{iter}}), e=1, ..., N^{iter}
\label{eq:anealing_schedule}
\end{equation}
where $p$ is the feature dimension of the original input data and $\mu$ is the annealing parameter which can be tuned using cross validation.
FSA has good computational efficiency and theoretical guarantees of consistency. The user defined parameter $k$ denoting how many features to select is more intuitive than the penalty parameter in the penalized methods (e.g. $L_1$ penalized regression) and makes the procedure more controllable.

\subsection*{Learning algorithm hyper-parameters}

Some learning algorithms such as FSA and boosted trees have their performance highly dependent on the values of the hyper-parameters.
To avoid any confounding effect of the method for selecting these parameters (e.g. by cross-validation or AIC/BIC), these learning algorithms were run on a discrete set of combinations on a single training/validation split of the data, and the parameter combination that obtained the best validation result was used in the entire experiment. 
The values that were used are given in Tables \ref{tab:paramTabFSA} and \ref{tab:paramTabBT}. 
The other learning algorithms were built-in Matlab and we used the default values for all parameters.

\begin{table}[htb]
\vspace{-3mm}
\begin{center}
\caption{Selected parameter values for FSA.}
\label{tab:paramTabFSA}
\begin{tabular}{|l|c|c|c|c|c|}
\hline
Parameters&BMI &Tumor &CoEPrA2006& Indoorloc &Wikiface\\
\hline
\hline
learning rate $\eta$ &0.00001&0.000003&0.0001&0.00001&0.00005\\
\hline
number of epochs $N^{iter}$&150&50&100&250&450\\
\hline
annealing parameter $\mu$&800&30&650&200&250\\
\hline
minibatch size&285&250&15&30&150\\
\hline
shrinkage parameter &0.0001&0.001&0.9&0.0001&0.001\\
\hline
\hline
Parameters&Gisette&Dexter&Madelon&SMK\_CAN\_187&GLI\_85\\
\hline
\hline
learning rate $\eta$ &0.0001&0.000001&0.0005&0.01&0.1\\
\hline
number of epochs $N^{iter}$ &60&300&10&500&800\\
\hline
annealing parameter $\mu$&600&100&40&280&100\\
\hline
minibatch size&20&30&145&145&100\\
\hline
shrinkage parameter&0&0&0.00001&0.001&0.005\\
\hline
\end{tabular}
\end{center}
\vspace{-5mm}
\end{table}

\begin{table}[htb]
\vspace{-3mm}
\begin{center}
\caption{Selected parameter values for boosted trees.}
\label{tab:paramTabBT}
\begin{tabular}{|l|c|c|c|c|c|}
\hline
Parameters&BMI &Tumor &CoEPrA2006& Indoorloc &Wikiface\\
\hline
\hline
max number of splits&1&1&1&8&1\\
\hline
boosting iterations&100&10&10&500&400\\
\hline
\hline
Parameters&Gisette&Dexter&Madelon&SMK\_CAN\_187&GLI\_85\\
\hline
\hline
max number of splits&4&4&$2^{6}$&1&2\\
\hline
boosting iterations&400&400&1900&500&200\\
\hline
\end{tabular}
\end{center}
\vspace{-5mm}
\end{table}

\subsection*{Table of groups}
In this section we present the summary of the performance of each screening method-learning algorithm combination and their division into groups such that the difference between the best method and the worst method in each group is not significant at the 0.05 level.

\begin{table}[ht]
\vspace{2mm}
\begin{center}
\caption{Table of groups, BMI dataset. SE is the standard error of mean estimation, $\omega$ is the number of features selected by the screening method, $\kappa$ is the number of features selected by FSA.}
\label{tab:grouptableBMI}
\begin{tabular}{|l|c|c|c|c|c|c|c|}
\hline
\multicolumn{2}{|c|}{Group} &{Screening Methods} &{Learner}&{Mean} &{SE} &{$\omega$} &{$\kappa$}\\
\hline
\hline
A && {RReliefF} &{FSA} &0.7632 &0.0006 &1758 &692 \\
\hline
A &&{---} &{FSA} &0.7607 &0.0005 &--- &839\\
\hline
A && {Mutual Information} &{FSA} &0.7606 &0.0006 &3537 &1550\\
\hline
A && {Correlation} &{FSA} &0.7590 &0.0006 &5856 &1354 \\
\hline
B && {Correlation} &{Ridge} &0.7238 &0.0008 &5140 &--- \\
\hline
B&C& {RReliefF} &{Ridge} &0.7172 &0.0005 &6230 &--- \\
\hline
D &C&{Mutual Information} &{Ridge}  &0.7078 &0.0009 &6230 &--- \\
\hline
D& &{---} &{Ridge}  &0.7073 &0.0004 &--- &--- \\
\hline
E& &{Mutual Information } &{Boosted Reg. Trees} &0.5198 &0.0020 &13 &--- \\
\hline
E& &{RReliefF } &{Boosted Reg. Trees} &0.5157 &0.0027 &13 &--- \\
\hline
E& &{Correlation } &{Boosted Reg. Trees} &0.4932 &0.0018 &13 &--- \\
\hline
F& &{--- } &{Boosted Reg. Trees} &0.2520 &0.0043 &--- &--- \\
\hline
\end{tabular}
\end{center}
\vspace{-6mm}
\end{table}

In Table \ref{tab:grouptableBMI} are shown the groups, the mean $R^2$ of test data and standard error of mean estimation obtained over all the runs for the BMI dataset. 
Also shown are the number of features $\omega$ selected by the screening method and the number of features $\kappa$ selected by the learning algorithm where the average $R^2$ is maximum. 
From Table \ref{tab:grouptableBMI} we see that the best learner is FSA and that the FSA results with and without screening methods belong to the same group indicating that the screening methods don't improve the performance of FSA significantly. 
For ridge regression, the performance of RReliefF and Mutual information belongs to a group higher than ridge regression without screening. For boosted regression trees, the screening methods do provide a significant improvement. 
We can also see that the number of features selected by FSA is smaller than the number of  features selected by the screening methods. 
So for FSA, the features selected by screening methods can still be reduced in order to get the best result.

\begin{table}[htb]
\vspace{2mm}
\begin{center}
\caption{Table of groups, Tumor dataset. SE is the standard error of mean estimation, $\omega$ is the number of features selected by the screening method, $\kappa$ is the number of features selected by FSA.}\label{tab:grouptabletumor}
\begin{tabular}{|l|c|c|c|c|c|c|c|}
\hline
\multicolumn{2}{|l|}{Group} &{Screening Methods} &{Learner}&{Mean} &{SE} &{$\omega$} &{$\kappa$}\\
\hline
\hline
A& &{---}&{FSA} &0.3473 &0.0001 &--- &558 \\
\hline
A&B&{RReliefF} &{ FSA} &0.3472 &0.0001 &6230 &2210 \\
\hline
C&B&{Correlation} &{FSA} &0.3427 &0.0001&6230 &1550 \\
\hline
C&&{Mutual Information} &{FSA} &0.3404 &0.0001 &6230 &1758 \\
\hline
D& &{Mutual Information} &{Ridge} &0.2949 &0.0001 &13 &---  \\
\hline
D& &{Correlation} &{ Ridge} &0.2925 &$<0.0001$ &13 &--- \\
\hline
D&E&{Correlation } &{ Boosted Reg. Trees} &0.2855 &0.0004 &13 &--- \\
\hline
D&E&{Mutual Information } &{ Boosted Reg. Trees} &0.2840 &0.0005 &13 &--- \\
\hline
&E&{RReliefF} &{ Ridge} &0.2831 &0.0001 &13 &--- \\
\hline
&E&{RReliefF } &{ Boosted Reg. Trees} &0.2738 &0.0003 &93 &--- \\
\hline
F& &{ --- } &{ Boosted Reg. Trees} &0.2272 &0.0003 &--- &---\\
\hline
F& &{---} &{ Ridge} &0.2153 &0.0003 &--- &--- \\
\hline
\end{tabular}
\end{center}
\vspace{-5mm}
\end{table}

The same types of results are shown in Table \ref{tab:grouptabletumor} for the tumor dataset. 
Again, the best results are obtained with FSA and the FSA results without screening methods belong to the first group. So screening methods do not improve the performance of FSA in this case either. 
For ridge regression and boosted regression trees, the results with screening methods belong to higher tier groups than results without screening method, which means the screening methods help  those two learners. 
Also for FSA, the number features selected by screening methods is further reduced in order to get the maximum result.

\begin{table}[htb]
\vspace{2mm}
\begin{center}
\caption{Table of groups, CoEPrA2006$\_$3 dataset. SE is the standard error of mean estimation, $\omega$ is the number of features selected by the screening method, $\kappa$ is the number of features selected by FSA.}\label{tab:grouptableCoEPrA20063}
\begin{tabular}{|l|c|c|c|c|c|c|c|c|}
\hline
\multicolumn{3}{|c|}{Group} &{Screening Methods} &{Learner}&{Mean} &{SE} &{$\omega$} &{$\kappa$}\\
\hline
\hline
A &&&{Correlation} &{Ridge} &0.2858 &0.0046 &208 &--- \\
\hline
A &B&&{---} &{FSA} &0.2844 &0.0049 &--- &65\\
\hline
A &B&&{RReliefF} &{FSA} &0.2815 &0.0061&2940 &65 \\
\hline
A &B &&{Correlation} &{FSA} &0.2747 &0.0044 &971 &412 \\
\hline
C&B &&{RReliefF} &{Ridge} &0.2482 &0.0052 &971 &--- \\
\hline
C&& &{Mutual Information} &{FSA} &0.2227 &0.0049 &3112 &65\\
\hline
D& &&{Mutual Information} &{Ridge}  &0.0763 &0.0053 &412 &--- \\
\hline
D&E&&{RReliefF } &{Boosted Reg. Trees} &0.0746 &0.0064 &491 &--- \\
\hline
D&E &F&{Correlation } &{Boosted Reg. Trees} &0.0661 &0.0064 &668 &--- \\
\hline
D&E&F&{--- } &{Boosted Reg. Trees} &0.0362 &0.0049 &--- &--- \\
\hline
&E&F&{Mutual Information } &{Boosted Reg. Trees} &0.0082 &0.0019 &65 &--- \\
\hline
&&F&{---} &{Ridge}  &0 &0 &--- &--- \\
\hline
\end{tabular}
\end{center}
\vspace{-6mm}
\end{table}

In Table \ref{tab:grouptableCoEPrA20063} are shown the results for the CoEPrA2006$\_$3 dataset. 
Again the FSA without screening is in the top group. 
The results with screening methods for Ridge regression belong to higher tier groups than without screening. 
The results of boosted regression trees with or without screening belong to the same group.
So in this case, the screening methods only improve the performance of ridge regression.

\begin{table}[htb]
\vspace{2mm}
\begin{center}
\caption{Table of groups, Indoorloc dataset. SE is the standard error of mean estimation, $\omega$ is the number of features selected by the screening method, $\kappa$ is the number of features selected by FSA.}\label{tab:grouptableindoor}
\begin{tabular}{|l|c|c|c|c|c|c|}
\hline
\multicolumn{1}{|c|}{Group} &{Screening Methods} &{Learner}&{Mean} &{SE}  &{$\omega$} &{$\kappa$}\\
\hline
\hline
A&{Mutual Information } &{Boosted Reg. Trees} &0.9703 &$<0.0001$ &254 &--- \\
\hline
A&{RReliefF } &{Boosted Reg. Trees} &0.9698 &$<0.0001$ &285 &--- \\
\hline
B&{--- } &{Boosted Reg. Trees} &0.9685 &$<0.0001$ &--- &--- \\
\hline
B&{Correlation } &{Boosted Reg. Trees} &0.9681 &$<0.0001$ &381 &--- \\
\hline
C&{Mutual Information} &{Ridge}  &0.9198 &$<0.0001$ &397 &--- \\
\hline
C&{---} &{Ridge}  &0.9198 &$<0.0001$ &--- &--- \\
\hline
D&{Correlation} &{Ridge} &0.9188 &$<0.0001$ &397 &--- \\
\hline
E&{---} &{FSA} &0.9182 &$<0.0001$ &--- &397\\
\hline
F&{Mutual Information} &{FSA} &0.9177 &$<0.0001$ &397 &285\\
\hline
G&{Correlation} &{FSA} &0.9167 &$<0.0001$ &397 &300 \\
\hline
H&{RReliefF} &{Ridge} &0.9158 &$<0.0001$ &397 &--- \\
\hline
I&{RReliefF} &{FSA} &0.9139 &$<0.0001$ &397 &300 \\
\hline
\end{tabular}
\end{center}
\vspace{-6mm}
\end{table}

In Table \ref{tab:grouptableindoor} are shown the results for the Indoorloc dataset. Here we see that two results with screening methods for Boosted Trees belong to higher tier group than without screening. 
There are no screening methods that give higher tier results than no screening for FSA and ridge regression.
\begin{table}[htb]
\vspace{2mm}
\begin{center}
\caption{Table of groups, Wikiface dataset. SE is the standard error of mean estimation, $\omega$ is the number of features selected by the screening method, $\kappa$ is the number of features selected by FSA.}\label{tab:grouptableface}
\begin{tabular}{|l|c|c|c|c|c|c|}
\hline
\multicolumn{1}{|c|}{Group} &{Screening Methods} &{Learner}&{Mean} &{SE}  &{$\omega$} &{$\kappa$}\\
\hline
\hline
A&{Mutual Information} &{Ridge}  &0.3490 &$<0.0001$ &1739 &--- \\
\hline
B&{RReliefF} &{Ridge} &0.3482 &$<0.0001$ &1739 &--- \\
\hline
C&{Correlation} &{Ridge} &0.3478 &$<0.0001$ &1739 &--- \\
\hline
D&{---} &{Ridge}  &0.3468 &$<0.0001$ &--- &--- \\
\hline
E&{Mutual Information} &{FSA} &0.3426 &$<0.0001$&1739 &370\\
\hline
E&{---} &{FSA} &0.3424 &$<0.0001$&--- &440\\
\hline
E&{RReliefF} &{FSA} &0.3424 &$<0.0001$ &1739 &440 \\
\hline
F&{Correlation} &{FSA} &0.3419 &$<0.0001$&1381 &370 \\
\hline
F&{Correlation } &{Boosted Reg. Trees} &0.2981 &$<0.0001$&31 &--- \\
\hline
G&{RReliefF } &{Boosted Reg. Trees} &0.2546 &$<0.0001$&10 &--- \\
\hline
G&{Mutual Information } &{Boosted Reg. Trees} &0.2517 &0.0003 &955 &--- \\
\hline
G&{--- } &{Boosted Reg. Trees} &0.2156 &0.0003&--- &--- \\
\hline
\end{tabular}
\end{center}
\vspace{-6mm}
\end{table}

In Table \ref{tab:grouptableface} are shown the results for the Wikiface dataset. 
All screening methods applied to ridge regression belong to higher tier groups than ridge regression without screening, whereas only the correlation method on boosted trees shows improvement for the other two learners.
\begin{table}[htb]
\vspace{-2mm}
\begin{center}
\caption{Table of groups, Dexter dataset. SE is the standard error of mean estimation,  $\omega$ is the number of features selected by the screening method, $\kappa$ is the number of features selected by FSA.}
\label{tab:grouptabledexter}
\begin{tabular}{|l|c|c|c|c|c|c|c|c|c|c|c|}
\hline
\multicolumn{6}{|c|}{Group} &{Screening Methods} &{Learner}&{Mean} &{SE} &{$\omega$} &{$\kappa$}\\
\hline
\hline
A&&&&&&{Mutual Information} &{Logistic Reg.} &0.9854 &$<0.0001$  &1463 &--- \\
\hline
A&&&&&&{Chi-square Score} &{Boosted Decision Trees} &0.9852 &$<0.0001$ &2662 &--- \\
\hline
A&B&&&&&{Chi-square Score} &{Logistic Reg.} &0.9851 &$<0.0001$  &1828 &--- \\
\hline
A&B&C&&&&{Gini Index} &{Logistic Reg.} &0.9850 &$<0.0001$  &1828 &--- \\
\hline
A&B&C&D&&&{Gini Index} &{Boosted Decision Trees} &0.9848 &$<0.0001$ &1828 &--- \\
\hline
A&B&C&D&&&{Mutual Information} &{Boosted Decision Trees} &0.9844 &$<0.0001$ &2892 &--- \\
\hline
A&B&C&D&&&{---} &{Boosted Decision Trees} &0.9841 &$<0.0001$ &--- &---\\
\hline
E&B&C&D&&&{Fisher Score} &{Logistic Reg.}&0.9839 &$<0.0001$  &2441 &---\\
\hline
E&B&C&D&F&&{Mutual Information} &{SVM} &0.9838 &$<0.0001$  &3893 &--- \\
\hline
E&&C&D&F&&{T-score} &{Logistic Reg.} &0.9838 &$<0.0001$  &2441 &--- \\
\hline
E&&C&D&F&&{MRMR} &{Logistic Reg.} &0.9837 &$<0.0001$  &2441 &--- \\
\hline
E&&C&D&F&&{MRMR} &{SVM} &0.9835 &$<0.0001$  &4164 &--- \\
\hline
E&G&C&D&F&&{T-score} &{Boosted Decision Trees} &0.9835 &$<0.0001$ &2892 &--- \\
\hline
E&G&C&D&F&&{Fisher Score} &{Boosted Decision Trees} &0.9835 &$<0.0001$ &2892 &--- \\
\hline
E&G&C&D&F&H&{MRMR} &{Boosted Decision Trees} &0.9835 &$<0.0001$ &3130 &--- \\
\hline
E&G&&D&F&H&{---} &{SVM} &0.9834 &$<0.0001$  &--- &--- \\
\hline
E&G&&D&F&H&{T-score} &{SVM} &0.9831 &$<0.0001$  &5023 &--- \\
\hline
E&G&I&&F&H&{Mutual Information} &{FSA} &0.9830 &$<0.0001$  &1828 &1828 \\
\hline
&G&I&J&F&H&{Chi-square Score} &{FSA} &0.9827 &$<0.0001$  &1828 &1828 \\
\hline
&G&I&J&F&H&{Gini Index} &{FSA} &0.9827 &$<0.0001$  &1828 &1828 \\
\hline
&G&I&J&&H&{T-score} &{FSA} &0.9824 &$<0.0001$  &5023 &5023 \\
\hline
&&I&J&&H&{MRMR} &{FSA} &0.9824 &$<0.0001$ &2662 &2441 \\
\hline
&&I&J&&H&{Fisher Score} &{FSA} &0.9823 &$<0.0001$  &2662 &2441 \\
\hline
&&I&J&&H&{Chi-square Score} &{SVM} &0.9822 &$<0.0001$ &3893 &--- \\
\hline
&&I&J&&H&{Gini Index} &{SVM} &0.9822 &$<0.0001$ &3893 &--- \\
\hline
&&I&J&&H&{---} &{Logistic Reg.}&0.9820 &$<0.0001$  &--- &---\\
\hline
&&I&J&&&{---}&{FSA}&0.9819 &$<0.0001$  &--- &2662 \\
\hline
K&&&J&&&{Fisher Score} &{SVM} &0.9809 &$<0.0001$  &3130 &--- \\
\hline
K&L&&&&&{Relief} &{Boosted Decision Trees} &0.9790 &$<0.0001$ &5023 &--- \\
\hline
&L&&&&&{Relief} &{FSA } &0.9780&$<0.0001$  &3130 &377 \\
\hline
&L&&&&&{Relief} &{Logistic Reg.}&0.9761 &0.0001  &1463 &---\\
\hline
M&&&&&&{Mutual Information} &{Naive Bayes} &0.9157 &0.0004 &83 &--- \\
\hline
N&&&&&&{MRMR} &{Naive Bayes} &0.9005 &0.0002 &83 &--- \\
\hline
N&&&&&&{Chi-square Score} &{Naive Bayes} &0.9002 &0.0003  &41 &---\\
\hline
N&&&&&&{Gini Index} &{Naive Bayes}&0.9002 &0.0003  &41 &--- \\
\hline
N&&&&&&{T-score} &{Naive Bayes} &0.8993 &0.0001  &83 &--- \\
\hline
N&&&&&&{Fisher Score} &{Naive Bayes}&0.8991 &0.0002  &83 &---\\
\hline
O&&&&&&{Relief} &{Naive Bayes}&0.8005 &0.0004  &12 &---\\
\hline
P&&&&&&{Relief} &{SVM} &0.6628 &0.0014  &41 &--- \\
\hline
P&&&&&&{---} &{Naive Bayes}&0.6520 &0.0006  &--- &---\\
\hline
\end{tabular}
\end{center}
\vspace{-6mm}
\end{table}

In Table \ref{tab:grouptabledexter} are shown the results for Dexter, a classification dataset. 
We see that for SVM, FSA and boosted trees the results of the learners with screening belong to either the same group or lower groups than learners without screening. 
Most of the screening methods did a great job in improving the performance of Logistic Regression for this dataset, and all methods improved the performance Naive Bayes. The Relief method didn't work on this data as all of the Relief based combinations are ranked at the end of table.
For some of the FSA combinations, the number of selected features by screening methods and number of selected features by FSA are the same, meaning the screening methods already selected the features that can give the best result. 

\begin{table}[htb]
\vspace{-3mm}
\begin{center}
\caption{Table of groups, Gisette dataset. SE is the standard error of mean estimation, $\omega$ is the number of features selected by the screening method, $\kappa$ is the number of features selected by FSA.}
\label{tab:grouptablegisette}
\begin{tabular}{|l|c|c|c|c|c|c|c|c|}
\hline
\multicolumn{3}{|c|}{Group} &{Screening Methods} &{Learner}&{Mean} &{SE} &{$\omega$} &{$\kappa$}\\
\hline
\hline
A&&&{MRMR} &{Boosted Decision Trees} &0.9978 &$<0.0001$  &2133 &--- \\
\hline
A&&&{T-score} &{Boosted Decision Trees} &0.9978 &$<0.0001$  &1634 &--- \\
\hline
A&&&{Fisher Score} &{Boosted Decision Trees}&0.9978 &$<0.0001$  &1634 &---\\
\hline
A&&&{Mutual Information} &{Boosted Decision Trees}&0.9977 &$<0.0001$ &1333 &---\\
\hline
A&&&{Gini Index} &{Boosted Decision Trees}&0.9977 &$<0.0001$&2884 &--- \\
\hline
A&B&&{Chi-square Score} &{Boosted Decision Trees}&0.9977 &$<0.0001$ &2312 &---\\
\hline
A&B&&{Chi-square Score} &{FSA}&0.9977 &$<0.0001$  &1634 &1634\\
\hline
A&B&&{Gini Index} &{FSA} &0.9977 &$<0.0001$  &1960 &1960 \\
\hline
A&B&&{Mutual Information} &{FSA}&0.9976 &$<0.0001$  &1480 &1480\\
\hline
A&B&&{T-score} &{FSA} &0.9976 &$<0.0001$  &1794 &1794 \\
\hline
A&B&&{Fisher Score} &{FSA} &0.9976 &$<0.0001$  &1960 &1960\\
\hline
A&B&&{MRMR} &{FSA} &0.9976 &$<0.0001$ &3954 &1058 \\
\hline
C&B&&{Relief} &{Boosted Decision Trees}&0.9975 &$<0.0001$  &1058 &---\\
\hline
C&&&{Relief} &{FSA } &0.9974 &$<0.0001$ &2687 &1480 \\
\hline
C&D&&{---}&{FSA}&0.9973 &$<0.0001$  &--- &1058\\
\hline
E&D&&{T-score} &{SVM} &0.9973 &$<0.0001$  &1480 &--- \\
\hline
E&D&&{Fisher Score} &{SVM}&0.9973 &$<0.0001$  &1634 &---\\
\hline
E&&&{Gini Index} &{SVM} &0.9972 &$<0.0001$ &1634 &--- \\
\hline
E&&&{MRMR} &{SVM}&0.9972 &$<0.0001$  &1960 &--- \\
\hline
E&F&&{Chi-square Score} &{SVM}&0.9972 &$<0.0001$  &1634 &--- \\
\hline
E&F&G&{Mutual Information} &{SVM}&0.9971 &$<0.0001$ &1480 &--- \\
\hline
H&F&G&{Mutual Information} &{Logistic Reg.}&0.9970 &$<0.0001$ &2133 &---\\
\hline
H&F&G&{---} &{Boosted Decision Trees} &0.9970 &$<0.0001$  &--- &--- \\
\hline
H&F&G&{Fisher Score} &{Logistic Reg.}&0.9969 &$<0.0001$  &1960 &--- \\
\hline
H&&G&{Chi-square Score} &{Logistic Reg.}&0.9969 &$<0.0001$  &1794 &---\\
\hline
H&&&{Gini Index} &{Logistic Reg.}&0.9969 &$<0.0001$  &1960 &--- \\
\hline
H&I&&{MRMR} &{Logistic Reg.}&0.9968 &$<0.0001$  &1794 &---\\
\hline
H&I&&{T-score} &{Logistic Reg.}&0.9968 &$<0.0001$ &1794 &--- \\
\hline
&I&&{Relief} &{Logistic Reg.} &0.9967 &$<0.0001$  &2497 &---\\
\hline
J&&&{---} &{SVM}&0.9963 &$<0.0001$  &--- &--- \\
\hline
K&&&{---} &{Logistic Reg.} &0.9962 &$<0.0001$ &--- &---\\
\hline
L&&&{Mutual Information} &{Naive Bayes}&0.9583 &$<0.0001$  &2312 &---\\
\hline
L&&&{MRMR} &{Naive Bayes}&0.9582 &$<0.0001$ &178 &--- \\
\hline
L&&&{T-score} &{Naive Bayes}&0.9582 &$<0.0001$  &1634 &---\\
\hline
L&&&{Fisher Score} &{Naive Bayes}&0.9582 &$<0.0001$  &2687 &---\\
\hline
L&&&{Gini Index} &{Naive Bayes}&0.9579 &$<0.0001$ &1794 &--- \\
\hline
L&&&{Chi-square Score} &{Naive Bayes}&0.9579 &$<0.0001$  &1794 &---\\
\hline
M&&&{Relief} &{Naive Bayes}&0.9474 &$<0.0001$  &2687 &---\\
\hline
N&&&{---} &{Naive Bayes}&0.9326 &$<0.0001$ &--- &---\\
\hline
O&&&{Relief} &{SVM}&0.8914 &$<0.0001$  &1333 &--- \\
\hline
\end{tabular}
\end{center}
\vspace{-5mm}
\end{table}

In Table \ref{tab:grouptablegisette} are shown the results for Gisette. Clearly screening methods work on boosted trees by giving results that belong to higher tier groups than the learner alone. 
Naive Bayes and logistic regression have a similar conclusion as boosted trees.
Beside the Relief-FSA combination, the other screening methods applied to FSA and SVM show improvement.
For some of the FSA combinations, the number of selected features by screening methods and number of selected features by FSA are the same, meaning the screening methods already selected the features that can give the best result. 
\begin{table}[htb]
\vspace{-3mm}
\begin{center}
\caption{Table of groups, SMK\_CAN\_187 dataset. SE is the standard error of mean estimation, $\omega$ is the number of features selected by the screening method, $\kappa$ is the number of features selected by FSA.}
\label{tab:grouptable187}
\begin{tabular}{|l|c|c|c|c|c|c|c|c|c|c|c|}
\hline
\multicolumn{6}{|c|}{Group} &{Screening Methods} &{Learner}&{Mean} &{SE} &{$\omega$} &{$\kappa$}\\ 
\hline
\hline
A&&&&&&{Relief} &{FSA}&0.8107 &0.0008  &4729 &139\\
\hline
A&&&&&&{Mutual Information} &{SVM}&0.81013 &0.0009  &5023 &---\\
\hline
A&B&&&&&{Mutual Information} &{FSA}&0.8072 &0.0013  &4729 &207\\
\hline
A&B&&&&&{Gini Index} &{SVM}&0.8039 &0.0009  &5023 &---\\
\hline
A&B&&&&&{MRMR} &{SVM} &0.8032 &0.0006  &5023 &--- \\
\hline
A&B&C&&&&{Chi-square Score} &{FSA}&0.8023 &0.0010  &1463 &83\\
\hline
A&B&C&D&&&{Relief} &{Boosted Decision Trees}&0.8019 &0.0008  &5023 &--- \\
\hline
A&B&C&D&E&&{---} &{SVM} &0.8015 &0.0009  &--- &--- \\
\hline
A&B&C&D&E&&{T-score} &{SVM}&0.8012 &0.0010  &5023 &---\\
\hline
&B&C&D&E&&{Fisher Score} &{SVM} &0.8002 &0.0008  &5023 &--- \\
\hline
&B&C&D&E&&{Chi-square Score} &{SVM}&0.7996 &0.0016  &5023 &---\\
\hline
F&B&C&D&E&&{Gini Index} &{FSA}&0.7991 &0.0018  &4442 &83\\
\hline
F&B&C&D&E&&{T-score } &{ Boosted Decision Trees} &0.7950 &0.0006  &2441 &---\\
\hline
F&B&C&D&E&&{MRMR} &{Boosted Decision Trees}&0.7940 &0.0010  &4442 &---\\
\hline
F&B&C&D&E&G&{---}&{FSA}&0.7932 &0.0013 &--- &83 \\
\hline
F&B&C&D&E&G&{--- } &{ Boosted Decision Trees} &0.7926 &0.0007  &--- &---\\
\hline
F&&C&D&E&G&{Fisher Score} &{Boosted Decision Trees}&0.7912 &0.0012  &711 &---\\
\hline
F&&&D&E&G&{Gini Index} &{Boosted Decision Trees}&0.7899 &0.0010  &2023 &---\\
\hline
F&&&&E&G&{Fisher Score} &{FSA}&0.7895 &0.0012  &139 &12\\
\hline
F&&&&E&G&{T-score} &{FSA}&0.7878 &0.0009  &139 &12 \\
\hline
F&H&&&&G&{MRMR} &{FSA}&0.7871 &0.0014  &3130 &139 \\
\hline
F&H&&&&G&{Mutual Information} &{Boosted Decision Trees}&0.7818 &0.0013  &4442 &--\\
\hline
F&H&&&&G&{Chi-square Score } &{Boosted Decision Trees}&0.7812 &0.0008  &5023 &--- \\
\hline
F&H&&&&G&{Relief} &{Logistic Reg.}&0.7804 &0.0013  &2023 &--- \\
\hline
&H&I&&&G&{Gini Index} &{Logistic Reg.}&0.7706 &0.0010  &377 &---\\
\hline
&H&I&&&&{Chi-square Score} &{Logistic Reg.} &0.7689 &0.0013  &286 &---\\
\hline
&&I&&&&{Fisher Score} &{Logistic Reg.}&0.7646 &0.0012  &589 &--- \\
\hline
&&I&&&&{T-score} &{Logistic Reg.}&0.7642 &0.0018  &842 &---\\
\hline
&&I&&&&{MRMR} &{Logistic Reg.}&0.7628 &0.0011  &711 &--- \\
\hline
&&I&&&&{Mutual Information} &{Logistic Reg.}&0.7567 &0.0017  &589 &---\\
\hline
J&&&&&&{MRMR} &{Naive Bayes} &0.7409 &0.0010  &12 &--- \\
\hline
J&K&&&&&{Fisher Score} &{Naive Bayes}&0.7312 &0.0008  &41 &--- \\
\hline
&K&L&&&&{T-score} &{Naive Bayes}&0.7291 &0.0008  &41 &---\\
\hline
&K&L&M&&&{Chi-square Score} &{Naive Bayes}&0.7279 &0.0009  &41 &---\\
\hline
&K&L&M&&&{Gini Index} &{Naive Bayes}&0.7249 &0.0005  &41 &--- \\
\hline
&&L&M&&&{Mutual Information} &{Naive Bayes}&0.7238 &0.0008  &41 &---\\
\hline
&&L&M&&&{---} &{Logistic Reg.}&0.7174 &0.0015  &--- &---\\
\hline
&&&M&&&{Relief} &{Naive Bayes}&0.7133 &0.0009  &12 &---\\
\hline
N&&&&&&{---} &{Naive Bayes}&0.6599 &0.0005  &--- &--- \\
\hline
O&&&&&&{Relief} &{SVM}&0.4730 &0.0006  &12 &---\\
\hline
\end{tabular}
\end{center}
\vspace{-5mm}
\end{table}

In Table \ref{tab:grouptable187} are shown the results for the SMK\_CAN\_187 dataset. 
The results with screening for Naive Bayes and Logistic Regression belong to higher tier groups than those without screening. 
For the other learning algorithms, screening methods give results belonging to the same group or lower groups as learners without screening. This indicates no improvement from using screening  for those learners.
\begin{table}[htb]
\vspace{-3mm}
\begin{center}
\caption{Table of groups, Madelon dataset. SE is the standard error of mean estimation, $\omega$ is the number of features selected by the screening method, $\kappa$ is the number of features selected by FSA.}
\label{tab:grouptablemadelon}
\begin{tabular}{|l|c|c|c|c|c|c|c|c|}
\hline
\multicolumn{3}{|c|}{Group} &{Screening Methods}&{Learner}&{Mean} &{SE} &{$\omega$} &{$\kappa$}\\ 
\hline
\hline
A&&&{Relief} &{Boosted Decision Trees}&0.9554 &$<0.0001$  &22 &--- \\
\hline
B&&&{T-score} &{Boosted Decision Trees}&0.9476 &$<0.0001$  &13 &--- \\
\hline
B&&&{Fisher Score} &{ Boosted Decision Trees} &0.9476 &$<0.0001$  &13 &---\\
\hline
C&&&{MRMR} &{ Boosted Decision Trees} &0.9460 &$<0.0001$  &13 &---\\
\hline
D&&&{Gini Index} &{ Boosted Decision Trees}&0.9398 &$<0.0001$  &13 &---\\
\hline
E&&&{Chi-square Score} &{ Boosted Decision Trees}&0.9376 &$<0.0001$  &13 &--- \\
\hline
F&&&{Mutual Information} &{Boosted Decision Trees}&0.9232 &$<0.0001$  &13 &---\\
\hline
G&&&{---} &{ Boosted Decision Trees}&0.8679 &$<0.0001$  &--- &---\\
\hline
H&&&{Relief} &{Naive Bayes}&0.6884 &$<0.0001$  &13 &---\\
\hline
I&&&{T-score} &{Naive Bayes}&0.6832 &$<0.0001$  &13 &---\\
\hline
I&&&{Fisher Score} &{Naive Bayes}&0.6832 &$<0.0001$  &13 &--- \\
\hline
I&J&&{Gini Index} &{Naive Bayes}&0.6821 &$<0.0001$  &22 &--- \\
\hline
&J&&{Mutual Information} &{Naive Bayes}&0.6818 &$<0.0001$  &13 &---\\
\hline
&J&&{MRMR} &{Naive Bayes}&0.6817 &$<0.0001$  22 &---\\
\hline
&J&&{Chi-square Score} &{Naive Bayes}&0.6815 &$<0.0001$  &22 &---\\
\hline
K&&&{Relief} &{FSA }&0.6394 &$<0.0001$  &42 &6\\
\hline
K&L&&{Relief} &{Logistic Reg.}&0.6389 &$<0.0001$  &6 &---\\
\hline
K&L&&{Mutual Information} &{SVM}&0.6386 &$<0.0001$  &6 &--- \\
\hline
K&L&M&{MRMR} &{FSA}&0.6384 &$<0.0001$  &364 &6\\
\hline
K&L&M&{T-score} &{FSA}&0.6384 &$<0.0001$  &364 &6\\
\hline
K&L&M&{Fisher Score} &{FSA}&0.6384 &$<0.0001$  &364 &6 \\
\hline
K&L&M&{Chi-square Score} &{FSA}&0.6384 &$<0.0001$  &364 &6 \\
\hline
K&L&M&{Gini Index} &{FSA}&0.6381 &$<0.0001$  &364 &6\\
\hline
&L&M&{Mutual Information} &{FSA}&0.6381 &$<0.0001$  &42 &6\\
\hline
&L&M&{T-score} &{SVM}&0.6381 &$<0.0001$  &6 &---\\
\hline
&L&M&{Fisher Score} &{SVM}&0.6381 &$<0.0001$  &6 &---\\
\hline
&L&M&{Chi-square Score} &{SVM}&0.6380 &$<0.0001$  &6 &---\\
\hline
&&M&{Mutual Information} &{Logistic Reg.}&0.6379 &$<0.0001$  &6 &---\\
\hline
&&M&{Gini Index} &{SVM}&0.6378 &$<0.0001$  &6 &---\\
\hline
&&M&{---}&{FSA}&0.6377 &$<0.0001$  &--- &6 \\
\hline
&&M&{MRMR} &{SVM}&0.6376 &$<0.0001$  &6 &---\\
\hline
&&M&{T-score} &{Logistic Reg.}&0.6373 &$<0.0001$  &6 &---\\
\hline
&&M&{Fisher Score} &{Logistic Reg.}&0.6373 &$<0.0001$ &6 &---\\
\hline
&&M&{Chi-square Score} &{Logistic Reg.}&0.6372 &$<0.0001$  &6 &---\\
\hline
&&M&{Gini Index} &{Logistic Reg.}&0.6368 &$<0.0001$  &6 &--- \\
\hline
&&M&{MRMR} &{Logistic Reg.}&0.6366 &$<0.0001$  &6 &---\\
\hline
&&M&{---} &{Naive Bayes}&0.6360 &$<0.0001$  &--- &--- \\
\hline
N&&&{Relief} &{SVM}&0.6120 &0.0001  &6 &--- \\
\hline
O&&&{---} &{Logistic Reg.}&0.5744 &0.0002  &--- &---\\
\hline
P&&&{---} &{SVM}&0.5455 &0.0002  &--- &---\\
\hline
\end{tabular}
\end{center}
\vspace{-5mm}
\end{table}

In Table \ref{tab:grouptablemadelon} are shown the results for Madelon. 
The results with screening  for Naive Bayes, SVM, Boosted Decision Trees and Logistic Regression belong to higher tier groups than those without screening. 
For FSA, only the result of Relief/FSA belongs to higher tier group than FSA without screening.
\begin{table*}[htb]
\vspace{-3mm}
\begin{center}
\caption{Table of groups, GLI\_85 dataset. SE is the standard error of mean estimation, $\omega$ is the number of features selected by the screening method, $\kappa$ is the number of features selected by FSA.}
\label{tab:grouptableGLI}
\begin{tabular}{|l|c|c|c|c|c|c|c|c|c|}
\hline
\multicolumn{4}{|c|}{Group} &{Screening Methods} &{Learner}&{Mean} &{SE} &{$\omega$} &{$\kappa$}\\
\hline
\hline
A&&&&{Mutual Information} &{SVM} &0.9639 &0.0014  &4164 &--- \\
\hline
A&&&&{---}&{FSA}&0.9627 &0.0006  &--- &842 \\
\hline
A&&&&{Relief} &{Logistic Reg.}&0.9616 &0.0009  &41 &---\\
\hline
A&B&&&{Relief} &{FSA } &0.9594&0.0010  &41 &41 \\
\hline
A&B&C&&{Fisher Score} &{SVM} &0.9587 &0.0010  &286 &--- \\
\hline
A&B&C&D&{Gini Index} &{FSA} &0.9560 &0.0014  &41 &12 \\
\hline
&B&C&D&{Mutual Information} &{FSA} &0.9555 &0.0005  &4729 &711 \\
\hline
&B&C&D&{---} &{SVM} &0.9548 &0.0007  &--- &--- \\
\hline
&B&C&D&{Gini Index} &{SVM} &0.9538 &0.0011  &3893 &--- \\
\hline
E&&C&D&{Fisher Score} &{FSA} &0.9529 &0.0012 &41 &12 \\
\hline
E&&C&D&{Chi-square Score} &{FSA} &0.9526 &0.0011  &41 &12 \\
\hline
E&&C&D&{Chi-square Score} &{SVM} &0.9524 &0.0009  &4442 &--- \\
\hline
E&F&C&D&{Mutual Information} &{Logistic Reg.} &0.9519 &0.0016  &4729 &--- \\
\hline
E&F&C&D&{Fisher Score} &{Logistic Reg.}&0.9516 &0.0008  &2892 &---\\
\hline
E&F&&D&{T-score} &{SVM} &0.9494 &0.0008  &4442 &--- \\
\hline
E&F&&D&{MRMR} &{FSA} &0.9492 &0.0008 &5023 &711 \\
\hline
E&F&&D&{MRMR} &{SVM} &0.9486 &0.0011  &3893 &--- \\
\hline
E&F&&D&{T-score} &{FSA} &0.9482 &0.0007  &5023 &711 \\
\hline
E&F&&D&{T-score} &{Logistic Reg.} &0.9478 &0.0008  &4729 &--- \\
\hline
E&F&G&D&{Chi-square Score} &{Logistic Reg.} &0.9450 &0.0017  &4729 &--- \\
\hline
E&F&G&&{Gini Index} &{Logistic Reg.} &0.9449 &0.0009  &2662 &--- \\
\hline
&F&G&&{MRMR} &{Logistic Reg.} &0.9445 &0.0006  &3893 &--- \\
\hline
H&F&G&&{Relief} &{Boosted Decision Trees} &0.9395 &0.0027  &139 &--- \\
\hline
H&F&G&&{Relief} &{Naive Bayes}&0.9393 &0.0007  &41 &---\\
\hline
H&F&G&&{Relief} &{SVM} &0.9379 &0.0013  &139 &--- \\
\hline
H&&G&&{Fisher Score} &{Boosted Decision Trees} &0.9357 &0.0012  &41 &--- \\
\hline
H&I&G&&{---} &{Logistic Reg.}&0.9308 &0.0023  &--- &---\\
\hline
H&I&G&&{Mutual Information} &{Boosted Decision Trees} &0.9285 &0.0016  &83 &--- \\
\hline
H&I&G&&{T-score} &{Boosted Decision Trees} &0.9265 &0.0024  &12 &--- \\
\hline
H&I&G&&{Gini Index} &{Boosted Decision Trees} &0.9249 &0.0019  &41 &--- \\
\hline
H&I&&&{MRMR} &{Boosted Decision Trees} &0.9243 &0.0016  &12 &--- \\
\hline
H&I&&&{Fisher Score} &{Naive Bayes}&0.9235 &0.0015  &12 &---\\
\hline
H&I&&&{Chi-square Score} &{Boosted Decision Trees} &0.9204 &0.0024 &41 &--- \\
\hline
H&I&J&&{Mutual Information} &{Naive Bayes} &0.9103 &0.0012  &41 &--- \\
\hline
&I&J&&{---} &{Boosted Decision Trees} &0.9072 &0.0020  &--- &---\\
\hline
K&&J&&{MRMR} &{Naive Bayes} &0.8940 &0.0011  &83 &--- \\
\hline
K&&J&&{T-score} &{Naive Bayes} &0.8936 &0.0013  &83 &--- \\
\hline
K&&&&{Chi-square Score} &{Naive Bayes} &0.8833 &0.0013  &41 &---\\
\hline
K&&&&{Gini Index} &{Naive Bayes}&0.8821 &0.0014  &41 &--- \\
\hline
L&&&&{---} &{Naive Bayes}&0.7006 &0.0033  &--- &---\\
\hline
\end{tabular}
\end{center}
\vspace{-5mm}
\end{table*}

In Table \ref{tab:grouptableGLI} are shown the results for the GLI\_85 dataset. 
The results with screening belong to the same group or lower groups than the learner alone for FSA.
SVM, Logistic Regression and boosted trees each have a few screening methods that give higher tier results. All screening methods give results belonging to higher groups than learner without screening for Naive Bayes.
\end{document}